\documentclass[final,5p,times,twocolumn]{elsarticle}

\usepackage{framed,multirow}
\usepackage{graphicx}
\usepackage{amssymb}
\usepackage{amsthm}

\usepackage{url}
\usepackage{xcolor}
\usepackage{hyperref}

\journal{Medical Image Analysis}

\begin{document}

\begin{frontmatter}


\title{CDDSA: Contrastive Domain Disentanglement and Style Augmentation for Generalizable Medical Image Segmentation}



%% \author{John Smith}
%% \address{California, United States}
\author[1]{Ran~Gu}
\author[1,9]{Guotai~Wang\corref{cor1}}
\cortext[cor1]{Corresponding author.}
\ead{Guotai.Wang@uestc.edu.cn}
% \fntext[fn1]{This is author footnote for second author.}
\author[1]{Jiangshan~Lu}
\author[2,3]{Jingyang~Zhang}
\author[4,9]{Wenhui~Lei}
\author[5,7]{Yinan~Chen}
\author[6]{Wenjun~Liao}
\author[6]{Shichuan~Zhang}
\author[7]{Kang~Li}
\author[8]{Dimitris~N.~Metaxas}
\author[1,5,9]{Shaoting~Zhang\corref{cor1}}
%% Third author's email
\ead{Zhangshaoting@uestc.edu.cn}

\address[1]{School of Mechanical and Electrical Engineering, University of Electronic Science and Technology of China, Chengdu, China}
\address[2]{School of Biomedical Engineering, Shanghai Jiao Tong University, Shanghai, China}
\address[3]{School of Biomedical Engineering, ShanghaiTech University, Shanghai, China.}
\address[4]{School of Electronic Information and Electrical Engineering, Shanghai Jiao Tong University, Shanghai, China}
\address[5]{Sensetime Research, Shanghai, China}
\address[6]{Department of Radiation Oncology, Sichuan Cancer Hospital and Institute, University of Electronic Science and Technology of China, Chengdu, China}
\address[7]{West China Hospital-SenseTime Joint Lab, West China Biomedical Big Data Center, Sichuan
University, Chengdu, China}
\address[8]{Department of Computer Science, Rutgers University, Piscataway NJ 08854, USA}
\address[9]{Shanghai AI Lab, Shanghai, China}

\begin{abstract}
Generalization to previously unseen images with potential domain shifts and different styles is essential for clinically applicable medical image segmentation, and the ability to disentangle domain-specific and domain-invariant features is key for achieving Domain  Generalization (DG). However, existing DG methods can hardly achieve effective disentanglement to get high generalizability. To deal with this problem, we propose an efficient Contrastive Domain Disentanglement and Style Augmentation (CDDSA) framework for generalizable medical image segmentation. First, a disentangle network is proposed to decompose an image into a domain-invariant anatomical representation and a domain-specific style code, where the former is sent to a segmentation model that is not affected by the domain shift, and the disentangle network is regularized by a decoder that combines the anatomical and style codes to reconstruct the input image. Second, to achieve better disentanglement, a contrastive loss is proposed to encourage the style codes from the same domain and different domains to be compact and divergent, respectively. Thirdly, to further improve generalizability, we propose a style augmentation method based on the disentanglement representation to synthesize images in various unseen styles with shared anatomical structures. Our method was validated on a public multi-site fundus image dataset for optic cup and disc segmentation and an in-house multi-site Nasopharyngeal Carcinoma Magnetic Resonance Image (NPC-MRI) dataset for nasopharynx Gross Tumor Volume (GTVnx) segmentation. Experimental results showed that the proposed CDDSA achieved remarkable generalizability across different domains, and it outperformed several state-of-the-art methods in domain-generalizable segmentation. Code is available at~\url{https://github.com/HiLab-git/DAG4MIA}
\end{abstract}


\end{frontmatter}

% \linenumbers

\section{Introduction}
\label{sec1:intro}
Deep learning with Convolutional Neural Networks (CNNs) has achieved remarkable performance in medical image segmentation~\citep{ronneberger2015u,shen2017deep,gu2021comprehensive}, and most existing models are built on the assumption that training and testing images are from the same domain and have very similar, if not the same, distributions. However, in clinical practice, this assumption does often not hold due to several factors such as the differences in scanning devices, imaging protocols, patient groups and image quality between training and testing images, where the testing images are usually acquired from a different medical center than the training set. Such differences (a.k.a domain shift~\citep{guan2021domain}) can substantially degrade the model's performance at test time~\citep{ganin2016domain,kamnitsas2017unsupervised}.

To address this problem, many Domain Adaptation (DA) methods have been explored to transfer knowledge from a set of labeled images in a source domain to images in a target domain~\citep{tzeng2017adversarial,wu2022fpl,gu2022contrastive}. However, the DA methods need to tune the model's parameters based on a set of images in the target domain, which is not only time-consuming but also impractical if the the target domain is not known in advance~\citep{chen2018semantic}.  What's more, the model needs to be adapted to each target domain respectively, and is faced with the problem of catastrophic forgetting on previous domains, which is not scalable when applied to a range of new unknown domains. 

In contrast to DA, Domain Generalization (DG) that encourages a model to be generalizable to unseen domains is more appealing  and efficient as it does not need to tune the model after training. Recently, domain generalization has attracted increasing attentions in the field of both computer vision~\citep{wang2022generalizing} and  medical image analysis~\citep{dou2019domain,gu2021domain}.
Existing DG methods mainly include image- and feature-based approaches. For image-based approaches, data augmentation has been widely used for improving the generalizability of a model, and \cite{zhang2020generalizing} proposed BigAug that uses a series of stacked transformations to augment the training images, with the assumption that the shift between source and target domains could be simulated through extensive data augmentation. However, the configuration of data augmentation requires empirical settings and could be data-specific. In contrast, feature-based methods mainly focus on representation learning to extract most representative features for better  generalization across domains~\citep{wang2020dofe, gu2021domain}. \cite{wang2020dofe} introduced a domain-oriented feature embedding method that dynamically  enriches image features with domain prior knowledge learned from multi-site domains to make the semantic features more discriminative. \cite{gu2021domain} developed a Domain Composition and Attention Network (DCA-Net) that represents features in a certain domain as a linear combination of a set of basis representations in a representation bank, where the combination coefficients are obtained by an attention module. Both methods rely on a domain knowledge pool or a representation bank to infer the domain-specific knowledge to make the network aware of the domain of an input image, which helps to improve the generalizability. However, they are limited by the capacity and representation power of the knowledge pool/bank, and have a limited ability to recognize the invariant features across different domains.

\begin{figure*}
    \centering
    \includegraphics[width=0.68\textwidth]{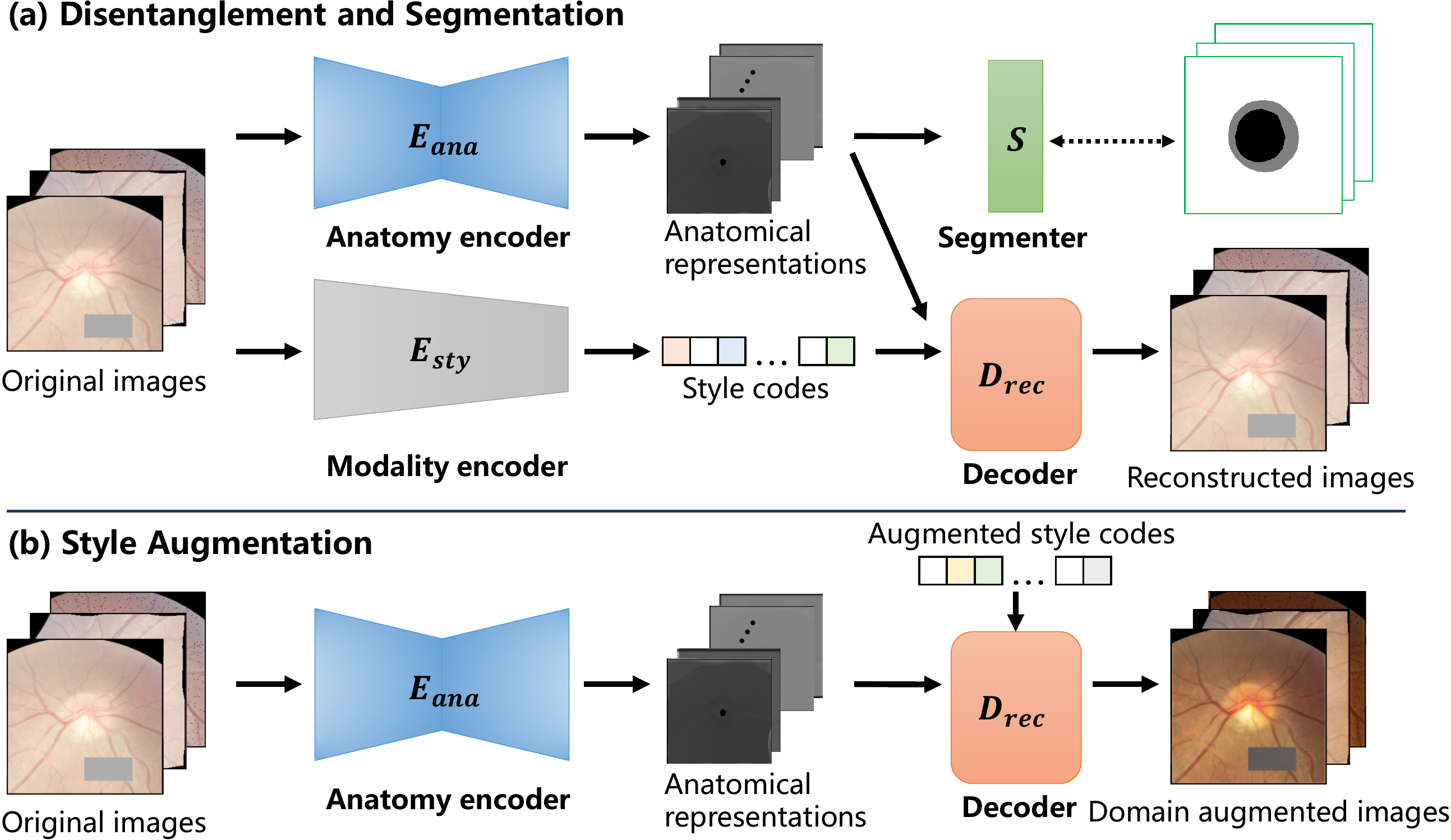}
    \caption{Workflow of our proposed Contrastive Domain Disentanglement and Style Augmentation (CDDSA) method. (a) shows the disentanglement and segmentation networks, where an anatomy encoder and a style encoder obtains anatomical representations and style codes respectively, and they are regularized by a decoder to reconstruct the input image. The segmentor takes domain-invariant anatomical representations as input to obtain the segmentation results. (b) represents the style augmentation strategy, where we  combine anatomical representations from a given image with augmented style codes to generate images in a new domain.} 
    \label{fig1:framework}
\end{figure*}

Recently, disentanglement has been introduced to computer vision that aims to explicitly decompose  features into domain-invariant contents and domain-specific styles~\citep{tran2017disentangled}.
It has also been employed to learn domain-invariant features for domain adaptation on multi-modality medical image segmentation datasets. \cite{yang2019unsupervised} applied disentangled representations to unsupervised domain adaptation for liver segmentation. They decomposed the images from two domains into a shared domain-invariant content space and a domain-specific style space, and used representations in the content space for segmentation. \cite{pei2021disentangle} used disentangled domain-invariant and domain-specific features for cardiac image segmentation across two modalities, and introduced a zero-loss to enhance the disentanglement. However, most existing disentangling methods are based on Generative Adversarial Networks (GAN), where a content encoder and a style encoder need to be trained for each known modality/domain, and multiple discriminators are involved, leading to a complex training process. Despite their suitability for domain adaptation, the GAN-based disentanglement methods are not scalable, as the number of required content/style encoders and discriminators will increase with the grow of domain number. What's more, such a paradigm cannot be applied to unseen domains as it requires the encoders for each domain to be trained in advance.  Therefore, they are not applicable to DG problems. 

In this work, we propose a novel GAN-free disentanglement framework named as Contrastive Domain Disentanglement and Style Augmentation (CDDSA) for domain-generalizable medical image segmentation. As shown in Fig.~\ref{fig1:framework}, it decomposes medical images in different domains into domain-invariant anatomical representations and domain-specific style codes with only one pair of anatomy Encoder and style encoder, which is regularized by a decoder that accepts an anatomical representation and  a style code to reconstruct an image. The encoders and decoder are  shared across different domains, without adversarial learning and domain-specific training, which is efficient and scalable to multiple domains. Our method was inspired by Spatial Decomposition Network (SDNet)~\citep{chartsias2019disentangled} that implements feature disentanglement without GAN.  Note that SDNet~\citep{chartsias2019disentangled} was proposed for semi-supervised learning, modality transformation and multi-modal image segmentation, and it can only perform disentanglement and image  reconstruction on seen domains with poor generalizability in unseen domains. The main reason is that SDNet lacks effective constraints on the style codes to encourage them to be domain-specific, which limits the ability to extract domain-invariant feature representations. In addition, it restricts the anatomical representations as binary codes, leading to a limited representation ability for effective image reconstruction. 

Differently from SDNet~\citep{chartsias2019disentangled}, our CDDSA is proposed for domain-generalizable segmentation of medical images. To improve the disentanglement performance, we relax the anatomical representation to soft values and propose a domain style contrastive learning loss to encourage the style codes in different domains to be discriminative from each other, which improves the model's ability to recognize domain-invariant anatomical representations that is sent to a segmentor to obtain segmentation results. As the segmentor is not affected by domain-specific features, it has a high generalizability across different domains. 
In addition, based on the extracted style codes in training domains, we can generate a new random style code and combine it with an existing anatomical representation to simulate images in an unseen domain with a new styles using the decoder, i.e., style augmentation, which further improves the generalizability of our framework.   

To the best of our knowledge, this is the first work in the literature to propose feature disentanglement learning for domain-generalizable medical image segmentation. The contributions of our method are summarised as follows:
\begin{itemize}
\item[1)] We introduce a novel framework CDDSA using GAN-free disentanglement for domain generalization in medical image segmentation. It achieves generalizability by segmentation from decomposed domain-invariant representations extracted by a single anatomy Encoder that is shared across domains and more efficient and scalable than GAN-based disentanglement.

\item[2)] To make the disentangled domain-specific style codes more representative and distinguishable, we propose domain style contrasitve learning, which forces the style codes from the same domain and different domains to be similar and dissimilar, respectively.

\item[3)] We propose style augmentation based on the disentangled anatomical representations and style codes to simulate images from unseen domains with different styles, which further improves the generalizability of the disentanglement and segmentation models. 

\item[4)] Comprehensive experimental results on multi-domain fundus images and multi-domain nasopharyngeal carcinoma magnetic resonance images (NPC-MRI) showed that our proposed CDDSA achieved high generalization on unseen domains, and it outperformed several state-of-the-art domain generalization methods.
\end{itemize}

\section{Related Works}
\label{sec2:relate_work}
\subsection{Domain Generalization for Medical Image Analysis}
Recently, domain generalization  has attracted increasing attentions to avoid dramatic performance degradation when inferring with images from unseen domains~\citep{li2018domain}. It aims to learn a model from a single or multiple source domains to make it directly applicable for unseen target domains without extra training~\citep{li2018learning,dou2019domain,wang2022generalizing}. Existing DG methods mainly include meta-learning methods, data-based methods and feature-based methods. Meta-learning~\citep{liu2020shape,dou2019domain} splits a set of source domains into meta-train and meta-test subsets, and adopts meta-optimization that iteratively updates model parameters to improve performance on the meta-test subset to simulate the situation when inferring on unseen domains.  \cite{liu2021feddg} combined meta-learning with federated learning to achieve privacy-preserving generalizable segmentation through continuous frequency space interpolation  across clients. However, meta-optimization process is highly time-consuming since all potential splitting results of meta-train and meta-test should be considered during training~\citep{dou2019domain}.

Data-based approaches usually use different data augmentation strategies for improving the model's generalizability. \cite{zhang2020generalizing} a deep stacked transformation  assuming that the shift between different domains can be simulated by  extensive data augmentation on a single domain. \cite{fick2021domain} utilized Cycle-GAN~\citep{zhu2017unpaired} to transform images from one certain domain to  other domains for augmentation. \cite{li2022domain} proposed Mixed Task Sampling (MTS) to enhance the variety of task-level training samples. Mixup in frequency domains~\citep{liu2021feddg,zhou2022ram_dsir} has  also  been used  to synthesize new images for model generalization. However, the efficiency of data augmentation largely depends on the ability to cover the data distribution in unseen domains, hence requiring empirical settings and even data-specific modifications.

Feature-based approaches  use domain-adaptive feature calibration or learn domain-invariant features to deal with domain generalization~\citep{wang2020dofe,muandet2013domain,liya2018domain}. % ben2006analysis
\cite{wang2020dofe} introduced a domain-oriented feature embedding framework that dynamically updates the domain-specific prior knowledge to make the semantic features more discriminative. \cite{hu2021domain} proposed a dynamic convolutional head to make the model's convolutional parameters adaptive to unseen target domains. \cite{gu2021domain} proposed a domain composition and attention method that calibrates the input feature based on attention coefficients represented by a representation bank. However, these methods did not explicitly obtain domain-invariant features for domain generalization, and they did not  separate features into purely domain-specific and domain-invariant representations well, leading to limited performance on domain generalization.

\subsection{Disentanglement Representation Learning}
Disentanglement explicitly decomposes features into domain-invariant contents and domain-specific styles~\citep{bengio2013representation,gatys2016image}. In addition to applications such as image synthesis~\citep{chartsias2019disentangled}, artifact removal and multi-task learning~\citep{meng2019representation} for medical image analysis,  it is widely adopted for domain adaptation~\citep{yang2019unsupervised,pei2021disentangle}. 
 \cite{yang2019unsupervised} used disentanglement to obtain domain-invariant content features for liver segmentation with domain adaptation. \cite{xie2020mi} used disentanglement to improve the performance of image translation for domain adaptation, and they disentangled the content features from domain information for both the source and translated images.
\cite{pei2021disentangle} applied disentanglement-based domain adaptation for cardiac image segmentation, and  introduced a zero loss to enhance disentanglement. \cite{ning2021new} 
proposed a bidirectional unsupervised DA framework based on disentangled representation learning for equally competent two-way DA performances on cardiac image segmentation. Despite their good performance on DA, theses works achieve disentanglement based on GAN, where multiple discriminators are needed in the adversarial training process that is complex and tricky to optimize. What's more, they need to have access to images for target domains during training, and are not applicable to DG tasks that involves unseen domains. \cite{chartsias2019disentangled} proposed a GAN-free Spatial Decomposition Network (SDNet) that decomposes an input image into a spatial factor (anatomy) and a non-spatial factor (style), and applied it to semi-supervised segmentation and image synthesis. However, it performs disentanglement and reconstruction well only on seen domains can hardly deal with unseen domains that are not involved in training.

\subsection{Contrastive Learning}
Contrastive learning is a self-supervised learning method to learn feature representations by enforcing positive pairs to be close and negative pairs to be distant~\citep{hadsell2006dimensionality}. Previous contrastive learning methods were mainly proposed to pre-train a powerful and representational feature extractor that can distinguish similar and dissimilar samples~\citep{he2020momentum,chen2020simple}. 
For computer vision and medical image analysis, contrastive learning has been mainly used for annotation-efficient learning. For example, \cite{kang2019contrastive} proposed a contrastive adaptation network that minimizes the intra-class domain discrepancy and maximizes the inter-class domain discrepancy.
\cite{chaitanya2020contrastive} used contrastive learning of global and local features sequentially for 3D medical image segmentation with limited annotations. \cite{lei2021contrastive} proposed contrastive learning of relative position regression for one-shot object localization in 3D medical images. \cite{You2022Miccai} proposed a contrastive voxel-wise representation learning to effectively learn low-level and high-level features for semi-supervised medical image segmentation.
Unlike these works, we design a contrastive learning strategy to enhance disentanglement between domain-invariant and domain-specific features to deal with domain generalization problems. 

\begin{figure*}
    \centering
    \includegraphics[width=\textwidth]{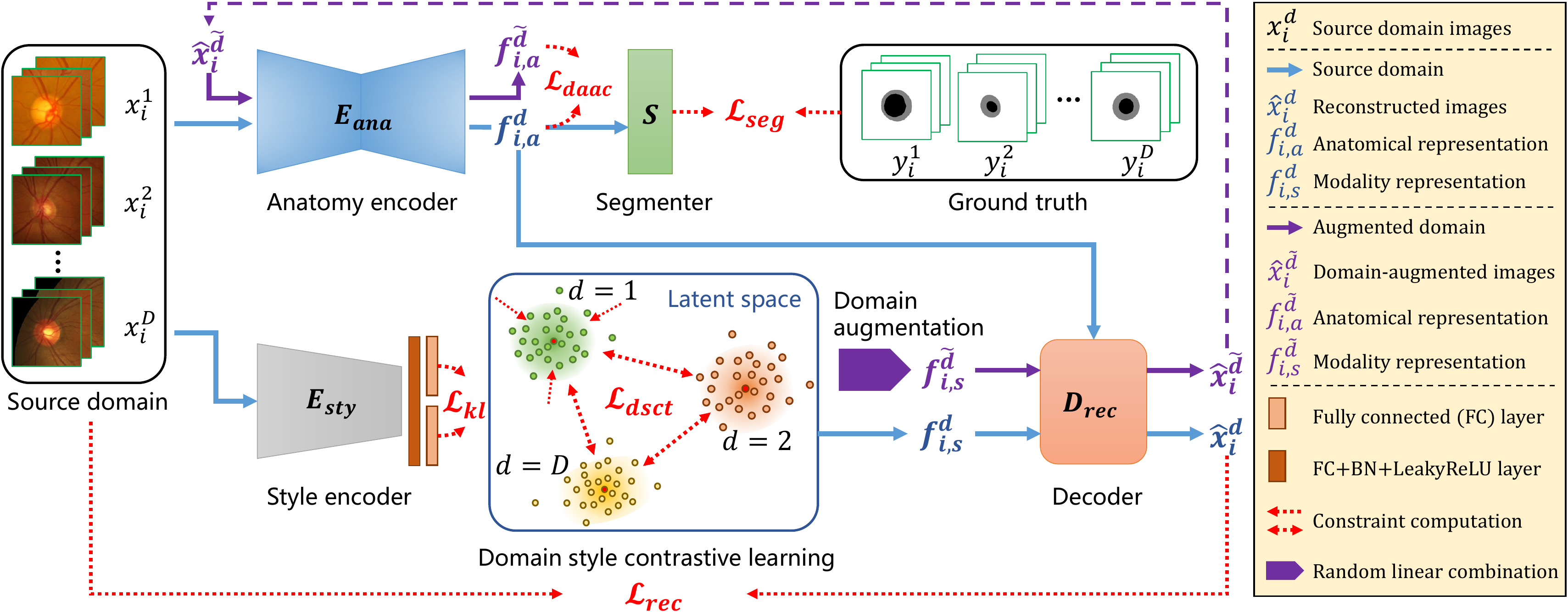}
    \caption{Overview of the proposed Contrastive Domain Disentanglement and Style Augmentation (CDDSA) network for multi-domain generalizable segmentation. We use an anatomy encoder $E_{ana}$ and a modality encoder $E_{sty}$ to extract anatomical representations $f_{a}^{d}$ and style codes $f_{s}^{d}$, respectively. A reconstruction decoder $D_{rec}$ takes $f_{a}^{d}$ and $f_{s}^{d}$ as input and obtains a reconstructed image $\hat{x}^{d}$. The decomposed anatomical representations $f_{a}^{d}$ is further used for segmentation. $\hat{x}^{\tilde{d}}$ is our simulated image using domain-augmentation strategy. $\hat{x}^{\tilde{d}}$ will further input into $E_{ana}$ to get its anatomical representation $f_{a}^{\tilde{d}}$. $\mathcal{L}_{saac}$ is used to encourage the consistency between $f_{a}^{\tilde{d}}$ and $f_{a}^{d}$.}
    \label{fig1:model}
\end{figure*}

\section{Methods}
\label{sec3:method}
For the domain generalization problem, the training set consists of images from $D$ domains and can be denoted as $\mathcal{D}=\{(x_i^d,y_i^d)\}_{i=1}^{N_d} ~(d=1,2,...,D)$, where $x_i^d$ depicts the $i$-th training sample from the $d$-th source domain with its corresponding ground-truth annotation $y_i^d$. $N_d$ denotes the number of training samples in domain $d$. 

Our proposed Contrastive Domain Disentanglement and Style Augmentation (CDDSA) framework is illustrated in Fig.~\ref{fig1:model}. Firstly, we employ a disentangle network containing an anatomy encoder $E_{ana}$ and a style encoder $E_{sty}$ to decompose an image into a domain-invariant anatomical representation and a domain-specific modality representation (i.e., style code), and they can be used to reconstruct the input image based on a decoder $D_{rec}$. 
We further send the disentangled anatomical representation into a segmenter $S$ to predict the segmentation mask. Secondly, to boost the disentanglement performance with more discriminative style codes across different domains, we introduce domain style contrasitve learning that forces the decomposed modality representations to have low intra-domain discrepancy and high inter-domain discrepancy. Thirdly, to further enhance model generalization, we proposed a style augmentation strategy to randomly generate style codes and combine them with given anatomical representations to reconstruct images with new styles that are not present in the training set. %We constrain the style codes in a smooth latent space through Variational Autoencoder (VAE)~\citep{kingma2013auto} so that the same distributed clusters in the space correspond to the similar domain distribution in the intensity space. Meanwhile, we save the disentangled anatomical representations as multi-channel feature maps, where each channel emphasizes different anatomical structures.

\subsection{Domain Disentangle Network.}
\label{sec3_1:domain_disentangle}
As shown in Fig.~\ref{fig1:model}, for an  input image $x_{i}^{d}$, we send it to an anatomy encoder $E_{ana}$ and a style encoder $E_{sty}$ to obtain an anatomical representation $f_{i,a}^d$ and  a modality representation (style code) $f_{i,s}^d$, respectively. Then  $f_{i,a}^d$ and $f_{i,s}^d$ are sent to a decoder $D_{rec}$ to reconstruct an input-like images $\hat{x}_{i}^d$, and a reconstruction loss $\mathcal{L}_{rec}$ is used to encourage the consistency between $x_{i}^{d}$ and $\hat{x}_{i}^d$. A segmentor $S$ takes $f_{i,a}^d$ as input to obtain the segmentation result.

\subsubsection{ Anatomy Encoder and Segmenter}
To decompose domain-invariant anatomical representations, we employ U-Net~\citep{ronneberger2015u} as the backbone to implement $E_{ana}$. We modify U-Net by setting the output channel of the last layer as $T$ and use tanh as the activation function in that layer. Let $H$ and $W$ represent the height and width of the input image $x_{i}^{d}$ respectively, the output of  $E_{ana}$ is denoted as $f_{i,a}^d\in [-1, 1]^{H\times W\times T}$, % that is expected to be domain-invariant, where $T$ is the number of channels for domain-invariant representation. $H$ and $W$ correspond to the height and width of the image, respectively. 
and we assume that each channel of $f_{i,a}^{d}$ emphasizes some anatomical information. Differently from SDNet~\citep{chartsias2019disentangled} that constrains $f_{i,a}^{d}$ to take binary values that may lose many details of object structures, we aim to reserve enough structural information for accurate image reconstruction and further style augmentation, and therefore soften the anatomical representation with a $tanh(x)= \frac{e^{x}-e^{-x}}{e^{x}+e^{-x}}$ activation function in the last layer. The anatomical representation extraction procedure is formulated as:
\begin{equation}
    \small
    f_{i,a}^d = E_{ana}(x_{i}^{d})
\end{equation}

Then, the decomposed anatomical representation $f_{i,a}^d$ is fed into a segmentation network $S$ to obtain a segmentation probability map $p^d_i = S(f^d_{i,a})$. Let $y^d_i$ denote the ground truth, and the supervised segmentation loss for domain $d$ is:
\begin{equation}
\small
\label{eq1:loss_seg}
    \mathcal{L}_{seg} = \frac{1}{2N_{d}}\sum_{i=1}^{N_{d}}\big( \mathcal{L}_{Dice}(p_i^{d},y_i^{d}) + \mathcal{L}_{ce}(p_i^{d},y_i^{d})\big)
\end{equation}
where we use a hybrid segmentation loss that consists of a Dice loss $\mathcal{L}_{Dice}$ and a cross-entropy loss $\mathcal{L}_{ce}$.

\subsubsection{Style Encoder}
The domain-specific modality representations are obtained by a style encoder $E_{sty}$ that is implemented by a Variational Autoencoder (VAE)~\citep{kingma2013auto}. The VAE learns a low dimensional latent space so that the learned latent representations match a prior distribution of an isotropic multivariate Gaussian $p(z)=\mathcal{N}(0, 1)$. Given the input $x_i^d$, $E_{sty}$ predicts the mean $u_{i}^d$ and variance $v_{i}^d$  of the distribution of a latent code $z \in \mathbb{R}^{1\times Z}$, where $Z$ is the length of the latent code. The style code $f_{i,s}^d$ of an input image  $x_{i,s}^d$ is sampled from the distribution characterized by mean $u_{i}^d$ and variance $v_{i}^d$. VAE is trained to minimize a reparameterization error, and a KL divergence loss is computed between the estimated Gaussian distribution $q(z|u_i^d, v_i^d)$ and the unit Gaussian $p(z)$:
\begin{equation}
\small
\label{eq3:loss_kl}
    \mathcal{L}_{kl} = D_{kl}\big(q(z|u_i^d,v_i^d) \| p(z)\big)
\end{equation}
where $D_{kl}(p\|q)=\sum p(x)log\frac{p(x)}{q(x)}$. When training is finished, sampling a vector from the unit Gaussian over a latent space can obtain a new style code, and we send it together  with an anatomical representation to the decoder to obtain a reconstructed image, where the decoder is used as a generative model, as detailed in the following. %passing it through the decoder approximates sampling from the input domain, and the decoder of VAE can be used as a generative model. %To obtain stable and effective domain-specific representations, the posterior distribution is modelled with a stochastic encoder as a convolutional network, which encodes the image modality as $f_{i,s}^{d}: x_{i}^{d}\rightarrow z$. Specifically, the stochasticity of the encoder is achieved as in the VAE formulation as $E_{sty}(x_{i}^d)$ produces first the mean and diagonal covariance for an $n_{z}$ dimensiona Gaussian, which is then sampled to yield the final $z$.

\subsubsection{Reconstruction Decoder}
\begin{figure}
    \centering
    \includegraphics[width=0.48\textwidth]{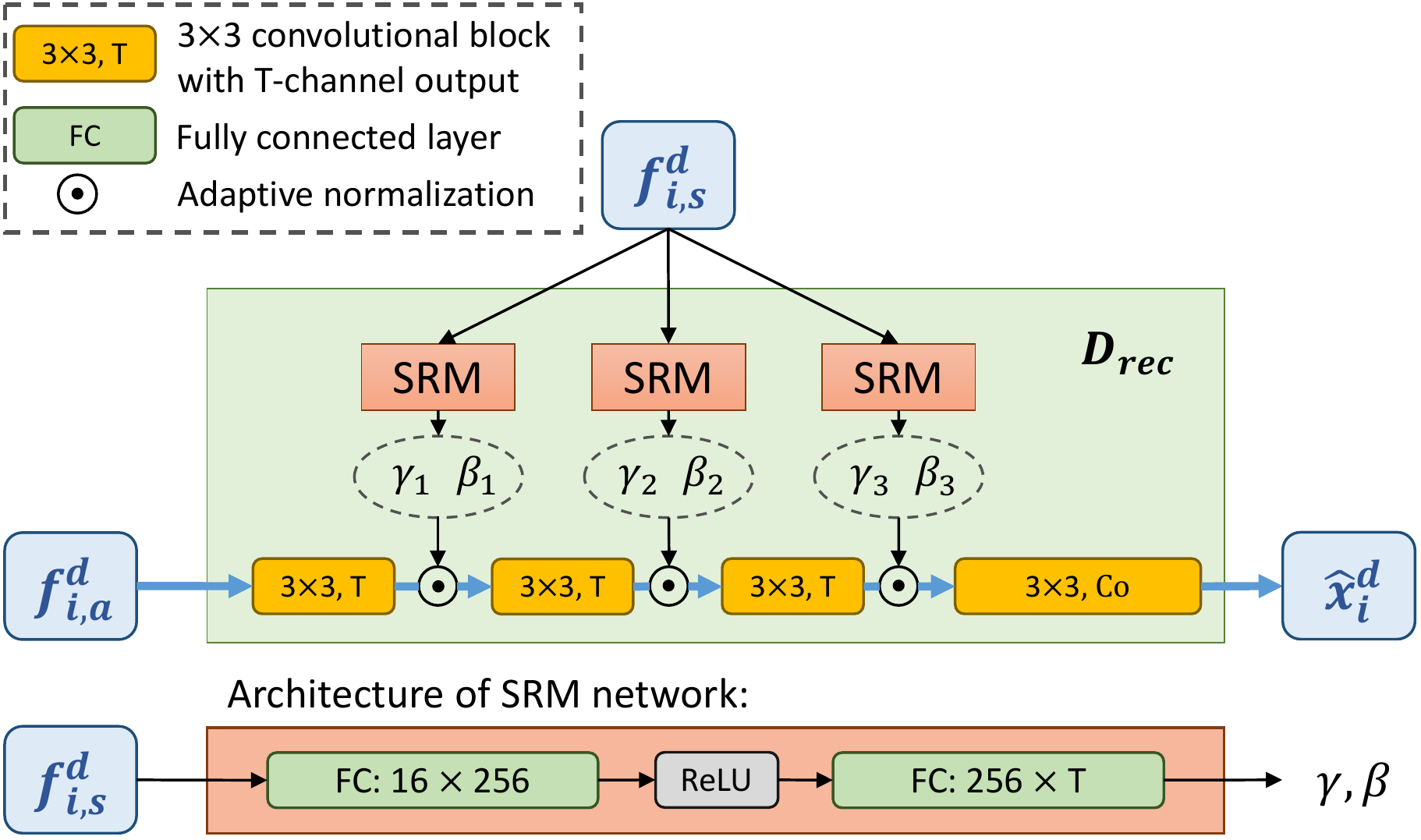}
    \caption{Framework of the reconstruction decoder $D_{rec}$. $T$ is the channel of feature maps and $Co$ represents the channel of output reconstructed image.} 
    \label{fig2-2:reconstruction_model}
\end{figure}

Fig.~\ref{fig2-2:reconstruction_model} shows the structure of our reconstruction decoder $D_{rec}$ to generate an image $\hat{x}_{i}^{d}$ given two decomposed representations $f_{i,a}^{d}$ and $f_{i,s}^{d}$. The collaboration of the two representations acts as a repainting mechanism where the anatomical representation $f_{i,a}^{d}$ is used to derive the anatomical content, and the modality representation $f_{i,s}^{d}$ is used to color the style distribution on the whole image~\citep{huang2018multimodal}. 

Specifically, the decoder uses four convolutional blocks to map $f_{i,a}^{d}$ to a reconstructed image conditioned on three Style Reconstruction Modules (SRM), as shown in Fig.~\ref{fig2-2:reconstruction_model}. For the intermediate feature map obtained by each convolutional block in the decoder, we apply Adaptive Instance Normalization (AdaIN) to control the output style, where the affine transformation parameters (scale and bias) are  predicted by an SRM that takes $f^d_{i,s}$ as input. 
%Each SRM learns the scale and bias parameters for each channel of a feature map within a convolutional architecture. Thus, an affine transformation learned from the conditioning input is applied that controlled Adaptive Instance Normalization (AdaIN) operating after each convolution layer of the reconstruction decoder~\citep{huang2017arbitrary}. 
Let $F_{i,c}$ represent the $c$-th channel of the intermediate feature map, we use two Fully Connected (FC) layers with a ReLU activation to implement the SRM that maps the modality representation $f_{i,s}^{d}$ to the scale $\gamma_{i,c}$ and bias $\beta_{i,c}$ that are used by affine transformation of AddIN:
\begin{equation}
    AdaIN(F_{i,c}|\gamma_{i,c}, \beta_{i,c}) =
    \gamma_{i,c}\frac{F_{i,c}-\mu(F_{i,c})}{\sigma(F_{i,c})} + \beta_{i,c}
\end{equation}
where each channel of the intermediate feature map is normalized separately, and we apply AddIN with SRM after each of three convolutional blocks in the decoder respectively, as shown in Fig.~\ref{fig2-2:reconstruction_model}. By mapping $f_{i,s}^{d}$ to the scale and bias values for each intermediate feature map, the reconstruction decoder $D_{rec}$ adaptively repaints style distribution on the anatomical representation $f_{i,a}^{d}$ in a coarse-to-fine manner. We use $\hat{x}_{i}^{d}$ to denoted the reconstructed image based on $f^d_{i,s}$ and $f^d_{i,a}$, and it is obtained by:  
\begin{equation}
    \hat{x}_{i}^{d} = D_{rec}(f_{i,s}^{d}, f_{i,a}^{d})
\end{equation}

As $f^d_{i,s}$ and $f^d_{i,a}$ are obtained from $x^d_i$, the reconstructed image $\hat{x}_{i}^{d}$ should be as close as possible to $x^d_i$. Therefore, a reconstruction loss is employed to train the  anatomy encoder $E_{ana}$, style encoder $E_{sty}$ and reconstruction decoder $D_{rec}$:
\begin{equation}
\small
\label{eq2:loss_rec}
    \mathcal{L}_{rec} = \frac{1}{N_{d}}\sum_{i=1}^{N_{d}}\big|x_i^d-\hat{x}_i^d\big|
\end{equation}
where we simply define the reconstruction loss as the Mean Absolute Error (MAE) loss  due to its robustness to outliers.

\subsection{Domain Style Contrastive Learning}
\label{sec3_2:style_contrastive}
An effective disentanglement expects that the style code $f_{i,s}^d$ to be domain-specific, but the reconstruction loss $\mathcal{L}_{rec}$ does not provide sufficient supervision for achieving domain-specific style codes. To address the problem 
%As shown in Fig.~\ref{fig1:model}, decomposing out two independent and discriminative representations ($f_{i,a}^{d}$ and $f_{i,s}^{d}$) and smoothly reconstructing the original-like input is crucial in disentanglement for self-supervised learning. However, traditional disentanglement only focuses on comprehensively disentangling in a specific image, which can be considered local discriminative disentanglement. In contrast, we expect that the disentangle model can summarize the global modality representations discriminatively in multi-site domains, leading these modality representations from the same domain to have the same distributions and those from different domains to be divergent from each other, while traditional local disentanglement ignores this issue. Hence, 
and make the model decompose more discriminative modality representations for different domains, we propose a domain style contrastive learning strategy to explictly constrain the disentangled style code $f_{i,s}^{d}$. 

Let $x_{i}^d$ and $x_{j}^d$ represent two different samples from the same domain $d$ in the training set, and their style codes obtained by $E_{sty}$ are denoted as $f_{i,s}^d$ and $f_{j,s}^d$, respectively. We define ($f_{i,s}^d$, $f_{j,s}^d$) as a positive pair for maximizing their similarity. At the same time, for $N$ samples each from a different domain $d'$ ($d' \in [0, 1, ..., D]$ and $d' \neq d$), their corresponding style codes compose a negative set $\mathcal{N}_i^d$ for $f_{i,s}^d$, and each element in $\mathcal{N}_i^d$ should have a minimized similarity compared with $f_{i,s}^d$.
%and $x_{n}^{d'}$ represents a training sample from a  domain other than $d$. 
%At each iteration, we store the disentangled modality representations of the specific domain $d$ together. 
Following the standard formula of self-supervised contrastive loss InfoNCE~\citep{oord2018representation,wang2021dense}, we define our domain style contrastive loss as:
\begin{equation}
\label{eq4:loss_sct}
\small
    \mathcal{L}_{dsct} = -log\frac{e^{sim(f_{i,s}^d, {f}_{j,s}^d)/\tau}}{e^{sim(f_{i,s}^d, {f}_{j,s}^d)/\tau}+\sum_{f\in\mathcal{N}_{i}^{d}} e^{sim(f_{i,s}^d, f)/\tau}}
\end{equation}
where $sim(\cdot,\cdot)$ is the cosine similarity, and $\tau=0.1$ is the temperature scaling parameter. In practice, to save the computational cost during training, we fetch $b$ samples for each domain in a mini-batch, and their style codes are saved in a list $Q$. Let $\tilde{Q}$ denote a permuted version of $Q$, the corresponding elements with the same index in the two lists are used as a positive pair, i.e., $b$ positive pairs are considered for each domain in a mini-batch. The style codes of the $b(D-1)$ samples from domains other than domain $d$ are used as the negative set for $x_{i}^d$.  

\subsection{Style Augmentation with Anatomical Consistency}
\label{sec3_3: domain-augmentation}
Based on the disentangled anatomical representations, style codes and the decoder, we can augment the style of an image by replacing its style code during image reconstruction, and therefore propose a style augmentation strategy to automatically generate images in new domains with different styles.  At each iteration of training, we denote the style codes of a batch as a style code bank $\mathcal{F} = \{f_{i, s}^d | i = 1, 2, ..., B; d = 1, 2, ..., D \}$, where the batch has $B$ samples for each domain. Based on the style codes in $\mathcal{F}$, we obtain a new style code using a linear combination of them with random weights: 
\begin{equation}
    f_{s}^{\tilde{d}} = \sum_{i=1}^{|\mathcal{F}|}\alpha_{i}\mathcal{F}_i
\end{equation}
where $f_{s}^{\tilde{d}}$ is a generated style code that is assumed to be from an unseen domain $\tilde{d}$. $\mathcal{F}_i$ is the $i$-th element in the style code bank, and the weight $\alpha_i \in [-1, 1]$ is randomly sampled from a uniform distribution.
%First, we store the decomposed modality representations from each domain in the training set at each iteration and construct a modality representation bank $f_{B,s}=\{f_{B,s}^{1}, ..., f_{B,s}^{k}\}$, where $B$ is the batch size, and therefore the number of the whole components in $f_{B,s}$ is $B\times k$. Since the proposed domain style contrastive learning encourages the model to decompose domain discriminative representations, it is desired to consider the items in $f_{B,s}$ as the typical representations in each domain, and an image can be summarized with a linear combination of these items~\citep{gu2021domain}. Hence, the proposed style augmentation strategy can obtain a new modality representation $f_s^{\tilde{d}}$ based on a linear combination with the random weights for modality representation bank $f_{B,s}$. Naturally, we generate random weight parameters $\alpha^d = [-1, 1]$ corresponding to the modality representations bank of each domain $d$. Smoothly, the newly simulated modality representations are formulated as:

Given an anatomical representation $f_{i,a}^d$ from an image in the source domain, we repaint it with the new style code $f_s^{\tilde{d}}$ to generate a new image $\hat{x}_{i}^{\tilde{d}}$:
\begin{equation}
\hat{x}_{i}^{\tilde{d}} = D_{rec}(f_{s}^{\tilde{d}}, f_{i,a}^{d})
\end{equation}

Since the generated image $\hat{x}_{i}^{\tilde{d}}$ and the real image $x_{i}^{d}$ share the same anatomical representation $f_{i,a}^d$, we introduce an anatomical consistency loss $\mathcal{L}_{saac}$ that forces the anatomy encoder $E_{ana}$ to obtain domain-invariant anatomical representations in spite of the different styles between $\hat{x}_{i}^{\tilde{d}}$ and $x_{i}^{d}$:
\begin{equation}
\small
\label{eq5:loss_dis}
    \mathcal{L}_{saac} = \frac{1}{ N_{d}}\sum_{i=1}^{N_{d}}\big|f_{i,a}^d-E_{ana}(\hat{x}_{i}^{\tilde{d}})|
\end{equation}
%perform a thorough decomposition ability and be robust on unseen-domain images, we feed the synthesized $\hat{x}_{i}^{\tilde{d}}$ back into $E_{ana}$ to obtain their corresponding anatomical representations $f_{i,a}^{\tilde{d}}$ that are expected to be consistent with $f_{i,a}^{d}$.
where the MAE loss is used for the anatomical consistency. 

\subsection{Overall Loss}
As a summary of the proposed CDDSA framework, the overall loss function for training is formulated as:
\begin{equation}
\small
\label{eq6:loss_all}
    \mathcal{L} = \mathcal{L}_{seg} + \lambda_{1}\mathcal{L}_{kl} + \lambda_{2}\mathcal{L}_{rec} + 
    \lambda_{3}\mathcal{L}_{dsct} +
    \lambda_{4}\mathcal{L}_{saac}
\end{equation}
where $\mathcal{L}_{seg}$ is the supervised segmentation loss (Eq.~\ref{eq1:loss_seg}), $\mathcal{L}_{kl}$ is the KL divergence loss for style encoder (Eq.~\ref{eq3:loss_kl}),  $\mathcal{L}_{rec}$ is the image reconstruction loss (Eq.~\ref{eq2:loss_rec}). $\mathcal{L}_{dsct}$ and  $\mathcal{L}_{saac}$ are the domain style contrastive loss (Eq.~\ref{eq4:loss_sct}) and style augmentation-based anatomical consistency loss (Eq.~\ref{eq5:loss_dis}), respectively.  %Losses from 1) and 2) are inherited from SDNet:
$\lambda_1,\lambda_2,\lambda_3$ and $\lambda_4$ act as trade-off parameters for different loss terms.
\begin{table}
    % \small
    \centering
    \caption{Statistics of retinal fundus images in four domains used in our experiment following~\cite{wang2020dofe}.} \label{tab0-1:fundus}
    \scalebox{0.8}{\begin{tabular}{lccc}
    \hline
    \hline
    Domain No. & Dataset & Cases (train / test) &  Scanner \\ \hline
    Domain 1 & Drishti-GS & 50 / 51 &  Aravind eye hospital \\
    Domain 2 & RIM-ONE-r3 & 99 / 60 &  Nidek AFC-210 \\
    Domain 3 & REFUGE-train & 320 / 80 &  Zeiss Visucam 500 \\
    Domain 4 & REFUGE-val & 320 / 80 &  Canon CR-2 \\ 
    \hline
    \hline
    % 99/60; 320/80; 320/80 & 789/281
    \end{tabular}}
\end{table}
\begin{table*}
    % \small
    \centering
    \caption{Statistics of the in-house multi-domain nasopharyngeal carcinoma  MRI dataset~\citep{liao2022automatic}}. %T1W means T1 weighted imaging. CE-T1W means  gadolinium-enhanced T1-weighted imaging. T1-water and T2-water represent T1 water imaging and T2 water imaging, respectively. SCPH means Sichuan Province People's Hospital. WCH represents West China Hospital.}
    \label{tab0-2:npc}
    \scalebox{0.75}{\begin{tabular}{lccccccc}
    \hline
    \hline
    Domain No. & Sequence & Slice thickness (mm) & Volumes (train / test) & Slices (train / test) & Total Volumes (train / test) & Total slices (train / test) & Scanner \\ \hline
    Domain 1 & T1 & 6 - 7.75 & 39 / 26 & 305 / 201 & \multirow{4}{*}{114 / 75} & \multirow{4}{*}{1427 / 994} & SPPH - Siemens \\
    Domain 2 & CE-T1 & 3 & 27 / 18 & 359 / 234 & & & WCH - Siemens \\
    Domain 3 & T1-water & 3 & 24 / 15 & 402 / 302 & & & WCH - Siemens \\
    Domain 4 & T2-water & 3 & 24 / 16 & 361 / 257 & & & WCH - Siemens \\
    \hline
    \hline
    % num: 39(305)/26(201); 27(359)/18(234); 24(402)/15(302); 24(361)/16(257) & 114(1427)/75(994) \\
    \end{tabular}}
\end{table*}

\section{Experiments and Results}
\subsection{Datasets and Implementation Details} 
In this study, we evaluated our proposed CDDSA and compared it with several state-of-the-art DG methods on a public multi-domain fundus image dataset and an in-house multi-domain  nasopharyngeal carcinoma MRI dataset.

\textbf{Multi-domain Fundus Image Dataset:} For a fair comparison with state-of-the-art DG methods, we evaluated our approach for Optic Cup (OC) and Disc (OD) segmentation on a public multi-domain retinal fundus image dataset~\footnote{https://github.com/emma-sjwang/Dofe}~\citep{wang2020dofe}. The dataset was collected from four public fundus image datasets obtained by different scanners at different sites that have distinct domain discrepancies in visual appearance and image quality: Domain 1 is from the Drishti-GS~\citep{sivaswamy2015comprehensive} dataset containing 50 and 51 images for training and testing, respectively; Domain 2 is from the RIM-ONE~\citep{fumero2011rim} dataset containing 99 and 60 images for training and testing, respectively; and the Domain 3 and 4 are from REFUGE~\citep{orlando2020refuge} challenge's training and validation datasets, respectively, and both of them contain 320 and 80 images for training and testing. 

To evaluate generalizability of OC$/$OD segmentation models in  unseen domains, we followed the leave-one-domain-out cross validation strategy  in DoFE~\citep{wang2020dofe}, where each time  three domains were used for training and the other  domain was used as the unseen testing domain. In total, there are 789 and  271 images for training and testing, respectively. The statistics of these multi-domain retinal fundus images are summarized in Table~\ref{tab0-1:fundus}. 
For preprocessing, we adopted a series of basic data augmentations to enhance the diversity of training samples as conducted by DoFE~\citep{wang2020dofe}, and the images were randomly cropped with a size of $256\times256$ during training.

\textbf{Multi-domain Nasopharyngeal Carcinoma MRI Dataset:} We collected an in-house multi-domain Nasopharyngeal Carcinoma (NPC) MRI dataset for nasopharynx Gross Tumor Volume (GTVnx) segmentation. It was collected from two hospitals with four different imaging protocols~\citep{liao2022automatic} (i.e., four domains): T1-wighted imaging, gadolinium contrast-enhanced T1-weighted (CE-T1) imaging, T1 water imaging and T2 water imaging, respectively. Images in Domain 1 were collected from Sichuan Provincial People's Hospital (SPPH) with  slice thickness of 6 - 7.75~mm, and images in Domain 2-4 were collected from  West China Hospital (WCH) with slice 
thickness of 3~mm.  In total, there were 189 volumes each from a specific patient, and they were split to 114 for training and 75 for testing.  The corresponding slice numbers for training and testing were 1427 and 994, respectively. The  volume and slice numbers for each  domain are detailed in Table~\ref{tab0-2:npc}. 

For preprocessing, we unified the orientation of different volumes into the standard RAI (right to left, anterior to posterior, inferior to superior in the x-, y-, and z-axes, respectively). The voxel intensity  was clipped by the 0.1 and 99.9 percentiles of each volume and then normalized to [0, 255]. Each volume was firstly cropped along z axis based on the slices containing GTVnx delineation, and then center-cropped with a $256 \times 256$ window in x-y plane. 
 We used 2D networks for the GTVnx segmentation in each slice and stacked the results into a 3D volume for evaluation, and  a leave-one-domain-out cross validation strategy was also employed during the experiment. 
%Sichuan Province People's Hospital containing 39 and 26 cases for training and testing, respectively. Domain 2 is axial 3D gadolinium-enhanced T1-weighted imaging with 3 mm slice thickness collected from West China Hospital containing 27 and 18 cases for training and testing, respectively. Domain 3 is T1 water imaging with 3 mm slice thickness from West China Hospital containing 24 and 15 cases for training and testing, respectively. Domain 4 is T2 water imaging with 3 mm slice thickness from West China Hospital containing 24 and 16 cases for training and testing, respectively. 
 %The statistics of in-house multi-site and sequence nasopharyngeal carcinoma datasets are summarized in Table.~\ref{tab0-1:fundus}.

\textbf{Implementation Details:} Training and inference were implemented on one NVIDIA GeForce GTX 1080 Ti GPU. The anatomical representation $E_{ana}$ was implemented by U-Net~\citep{ronneberger2015u} as the backbone, with channel numbers of 16, 32, 64, 128 and 256 at five resolution scales, respectively. We set the channel number of anatomical representations as $T=8$. The segmenter $S$ consists of two convolutional blocks. The first block has a convolution layer with a kernel size of  $3\times3$ followed by BN and  LeakyReLU (sloop $=0.2$), and the second 
block has a $1\times1$ convolution layer followed by Softmax to obtain a segmentation probability map.  The style encoder $E_{sty}$ has convolutional blocks each with a down-sampling layer to reduce the resolution, and the output of the last convolutional block is sent to two fully connected layers to obtain the mean and variance of a Gaussian distribution for the latent style code, and the size of the latent style code was set as $Z=16$. 

The weights in the total loss function were: $\lambda_{1} = 1.0$, $\lambda_{2}=0.001$, $\lambda_{3}=0.01$ and  $\lambda_{4}=1.0$ , respectively. 
The networks were trained with the Adam optimizer, and the learning rate was initialized to  $10^{-3}$ and decayed to 95\% when the performance did not improve in 8 epochs.
In a mini-batch, the image/slice number for each domain was $8$ and $6$ for the fundus image and NPC-MRI datasets, respectively. The epoch number was 200 and 400 for the fundus image and NPC-MRI datasets, respectively. To measure the segmentation performance quantitatively, we adopt the Dice score (Dice) and Average Symmetric Surface Distance (ASSD) for evaluation.

%For the fundus image dataset,  The proposed disentangle network was trained in an end-to-end manner using the Adam optimizer, and the initial learning rate was $1e^{-3}$. The learning rate will decay to 95\% when the model does not improve in 8 epochs. We trained 200 epochs with a batch size $B$ of 40 (10 for each domain), and the mini-batch size $b$ was 8 in each domain.  For fair comparison, we kept most parameters the same as those in DoFE. We trained the model in 400 epochs with a batch size $B$ of 24 (8 for each domain), and the mini-batch size $b$ was 6 in each domain. If there is no special explanation, the rest of the experimental settings are the same as those in fundus segmentation.
\begin{table*}
    % \small
    \centering
    \caption{Comparison of Dice (\%) by different DG methods on the multi-site fundus image dataset. %Baseline means the modified disentangle baseline. 
    %+$\mathcal{L}_{dsct}$ means baseline network introduces style contrastive learning. +$\mathcal{L}_{saac}$ means baseline network introduces style augmentation strategy. +$\mathcal{L}_{dsct}$+$\mathcal{L}_{saac}$ means baseline network introduces both style contrastive learning and style augmentation strategy that is our final proposed CDDSA model. 
    CDDSA$\diamond$ means      that the new style code for style augmentation was randomly sampled from a Gaussian distribution rather than obtained by  a random linear combination of style codes in the source domains.
    } \label{tab1:fundus_dice}
    \scalebox{0.76}{\begin{tabular}{l|c|c|c|c|c|c|c|c|c|c}
    \hline
    \hline
    \multirow{2}{0.74in}{Methods} & \multicolumn{2}{c}{Domain 1} & \multicolumn{2}{|c|}{Domain 2} & \multicolumn{2}{|c|}{Domain 3} & \multicolumn{2}{|c|}{Domain 4} &
    \multicolumn{2}{c}{Avg} \\ \cline{2-11}
    & cup & disc & cup & disc & cup & disc & cup & disc & cup & disc \\ \hline
    Lower bound (Inter-domain) & 74.38$\pm$12.96 & 96.67$\pm$2.04 & 77.71$\pm$20.84 & 85.05$\pm$14.67 & 79.72$\pm$9.51 & 90.01$\pm$5.81 & 86.63$\pm$8.52 & 89.55$\pm$3.26 & 79.61 & 90.32 \\
    Upper bound (Intra-domain) & 83.35$\pm$13.99 & 96.10$\pm$1.88 & 81.53$\pm$9.42 & 94.62$\pm$3.01 & 87.57$\pm$7.59 & 95.91$\pm$1.85 & 88.88$\pm$7.10 & 95.58$\pm$1.98 & 85.33 & 95.55 \\ \hline
    BigAug~\citep{zhang2020generalizing} & 82.36$\pm$11.74 & 93.73$\pm$9.29 & 75.45$\pm$15.01 & 87.83$\pm$11.17 &
    84.32$\pm$9.45 & 91.99$\pm$10.72 &
    85.32$\pm$7.50 & 92.97$\pm$6.58 & 81.86 & 91.63 \\
    DoFE~\citep{wang2020dofe} & 80.25$\pm$10.84 & 95.61$\pm$1.45 & 78.97$\pm$14.80 & 88.74$\pm$4.58 &
    84.81$\pm$7.71 & 92.81$\pm$2.63 & 86.65$\pm$6.39 & 93.46$\pm$2.43 & 82.67 & 92.66 \\
    FedDG~\citep{liu2021feddg} & 79.84$\pm$13.55 & 93.50$\pm$4.11 & 76.57$\pm$13.95 & 88.74$\pm$4.91 &
    84.23$\pm$6.80 & 93.73$\pm$3.22 & 85.33$\pm$10.19 & 94.03$\pm$4.14 & 
    81.49 & 92.50 \\
    DCA-Net~\citep{gu2021domain} & 
    82.16$\pm$12.23 & 94.39$\pm$2.94 & 
    80.63$\pm$15.58 & \textbf{91.50$\pm$2.78} &
    84.48$\pm$7.77 & 91.63$\pm$4.38 & 
    \textbf{87.11$\pm$12.67} & 93.05$\pm$4.98 & 
    83.60 & 92.64 \\ \hline
    %DCAC~\cite{hu2021domain} & 82.09$\pm$13.71 & 95.46$\pm$8.93 & 77.72$\pm$18.96 & 87.20$\pm$11.77 & 86.12$\pm$7.61 & 92.50$\pm$7.15 & 86.55$\pm$9.30 & 94.07$\pm$8.42 & 87.71 \\ \hline
    Baseline & 
    80.63$\pm$11.55 & 95.02$\pm$2.65 &
    79.35$\pm$13.66 & 89.76$\pm$3.20 &
    83.29$\pm$8.04 & \textbf{93.67$\pm$3.36} & 84.12$\pm$11.33 & 93.51$\pm$4.03 & 81.85 & 92.99 \\
    +$\mathcal{L}_{dsct}$ & 80.64$\pm$11.94 & 96.11$\pm$2.95 &
    80.13$\pm$17.39 & 88.27$\pm$1.23 &
    85.20$\pm$8.06 & 92.36$\pm$2.59 & 86.33$\pm$9.12 & 93.36$\pm$2.60 & 83.08 & 92.53 \\
    +$\mathcal{L}_{saac}$ & 84.28$\pm$11.56 & 96.13$\pm$1.35 &
    \textbf{81.95$\pm$12.60} & 88.18$\pm$3.50 &
    84.59$\pm$8.11 & 92.95$\pm$3.30 & 86.49$\pm$9.69 & 93.27$\pm$3.40 & 84.33 & 92.63 \\
    +$\mathcal{L}_{dsct}$+$\mathcal{L}_{saac}$ (CDDSA$\diamond$) & 85.53$\pm$11.37 & 96.74$\pm$1.70 & 
    76.39$\pm$17.45 & 88.25$\pm$6.91 &  84.60$\pm$7.76 & 92.34$\pm$4.08 & 86.56$\pm$9.75 & 92.63$\pm$3.39 & 83.27 & 92.49 \\
    +$\mathcal{L}_{dsct}$+$\mathcal{L}_{saac}$ (CDDSA) & \textbf{85.75$\pm$12.31} & \textbf{96.79$\pm$1.53} & 
    81.04$\pm$13.63 & 89.71$\pm$3.60 &  \textbf{86.94$\pm$7.94} & 93.25$\pm$3.55 & 86.86$\pm$8.97 & \textbf{94.44$\pm$3.96} & \textbf{85.15} & \textbf{93.55} \\
    \hline
    \hline
    \end{tabular}}
\end{table*}
\begin{table*}
    % \small
    \centering
    \caption{ Comparison of ASSD (pixel) by different DG methods on the multi-site fundus image dataset.}\label{tab2:fundus_asd}
    \scalebox{0.78}{\begin{tabular}{l|c|c|c|c|c|c|c|c|c|c}
    \hline
    \hline
    \multirow{2}{0.9in}{Methods} & \multicolumn{2}{c}{Domain 1} & \multicolumn{2}{|c|}{Domain 2} & \multicolumn{2}{|c|}{Domain 3} & \multicolumn{2}{|c|}{Domain 4} & \multicolumn{2}{|c}{Avg} \\ \cline{2-11}
    & cup & disc & cup & disc & cup & disc & cup & disc & cup & disc \\ \hline
    Lower bound (Inter-domain) & 22.35$\pm$9.74 & 6.47$\pm$3.80 & 15.77$\pm$20.21 & 18.25$\pm$19.60 & 12.30$\pm$5.82 & 12.33$\pm$5.03 & 7.45$\pm$4.60 & 9.27$\pm$2.62 & 14.47 & 11.58 \\
    Upper bound (Intra-domain) & 16.04$\pm$6.65 & 7.84$\pm$3.87 & 13.10$\pm$7.68 & 8.55$\pm$5.80 & 8.41$\pm$5.02 & 6.32$\pm$4.02 & 6.07$\pm$3.41 & 5.46$\pm$2.48 & 10.91 & 7.04 \\ \hline
    BigAug~\citep{zhang2020generalizing} & 17.91$\pm$10.11 & 8.67$\pm$4.08 & 22.33$\pm$15.26 & 19.77$\pm$6.69 &
    13.51$\pm$7.67 & 14.46$\pm$4.96 &
    8.90$\pm$5.02 & 8.77$\pm$6.63 & 15.66 & 12.92 \\
    DoFE~\citep{wang2020dofe} & 
    17.16$\pm$9.40 & 7.62$\pm$2.38 & 15.28$\pm$12.94 & 14.52$\pm$5.36 &
    10.73$\pm$6.22 & 10.11$\pm$5.11 & \textbf{7.18$\pm$3.23} & 7.60$\pm$3.64 & 12.59 & 9.96 \\ 
    FedDG~\citep{liu2021feddg} & 
    18.97$\pm$12.82 & 7.83$\pm$3.11 & 15.34$\pm$9.33 & 13.74$\pm$6.79 & 12.21$\pm$5.57 & 9.71$\pm$5.63 & 9.21$\pm$6.62 & 8.15$\pm$5.89 & 
    13.93 & 9.86 \\ 
    DCA-Net~\citep{gu2021domain} & 
    17.19$\pm$7.64 & 9.32$\pm$4.70 & 
    12.39$\pm$12.32 & 10.46$\pm$3.23 &
    11.28$\pm$5.21 & 11.32$\pm$5.54 & 
    7.37$\pm$6.51 & 7.22$\pm$4.75 & 
    12.06 & 9.58 \\ \hline
    %DCAC~\citep{hu2021domain} & 18.97$\pm$11.83 & 7.83$\pm$3.11 & 16.98$\pm$12.75 & 14.15$\pm$5.17 & 10.25$\pm$6.20 & 9.64$\pm$3.42 & 8.87$\pm$4.18 & \textbf{6.47$\pm$3.52} & 13.77 & 9.92 \\ \hline
    Baseline & 
    17.33$\pm$8.58 & 8.21$\pm$4.61 &
    13.10$\pm$6.66 & \textbf{11.79$\pm$3.90} & 11.03$\pm$5.22 & \textbf{9.31$\pm$4.23} & 8.04$\pm$6.42 & 7.43$\pm$4.89 & 
    12.38 & 9.18 \\
    +$\mathcal{L}_{dsct}$ & 
    18.21$\pm$8.05 & 7.52$\pm$5.85 &
    13.33$\pm$13.43 & 14.10$\pm$13.50 &
    10.09$\pm$5.42 & 9.98$\pm$3.11 & 7.30$\pm$3.88 & 7.03$\pm$2.57 & 
    12.23 & 9.66 \\
    +$\mathcal{L}_{saac}$ & 
    15.77$\pm$6.93 & 7.14$\pm$2.59 &
    \textbf{11.04$\pm$5.91} & 12.97$\pm$3.94 &
    10.58$\pm$5.21 & 9.60$\pm$3.71 & 7.22$\pm$5.31 & 7.51$\pm$4.10 & 
    11.15 & 9.31 \\
    +$\mathcal{L}_{dsct}$+$\mathcal{L}_{saac}$ (CDDSA$\diamond$) & 14.85$\pm$6.87 & 6.78$\pm$3.47 & 15.35$\pm$12.98 & 14.61$\pm$12.51 & 10.72$\pm$5.25 & 9.92$\pm$4.07 & 7.35$\pm$4.44 & 7.75$\pm$3.77 & 12.07 & 9.77 \\
    +$\mathcal{L}_{dsct}$+$\mathcal{L}_{saac}$ (CDDSA) & \textbf{14.65$\pm$8.39} & \textbf{6.54$\pm$3.74} & 12.91$\pm$10.79 & 13.06$\pm$8.60 &  \textbf{9.38$\pm$5.40} & 9.32$\pm$4.11 & 7.28$\pm$5.85 & 6.87$\pm$5.03 & \textbf{11.06} & \textbf{8.95} \\
    \hline
    \hline
    \end{tabular}}
\end{table*}
\begin{figure*}
    \centering
    \includegraphics[width=0.76\textwidth]{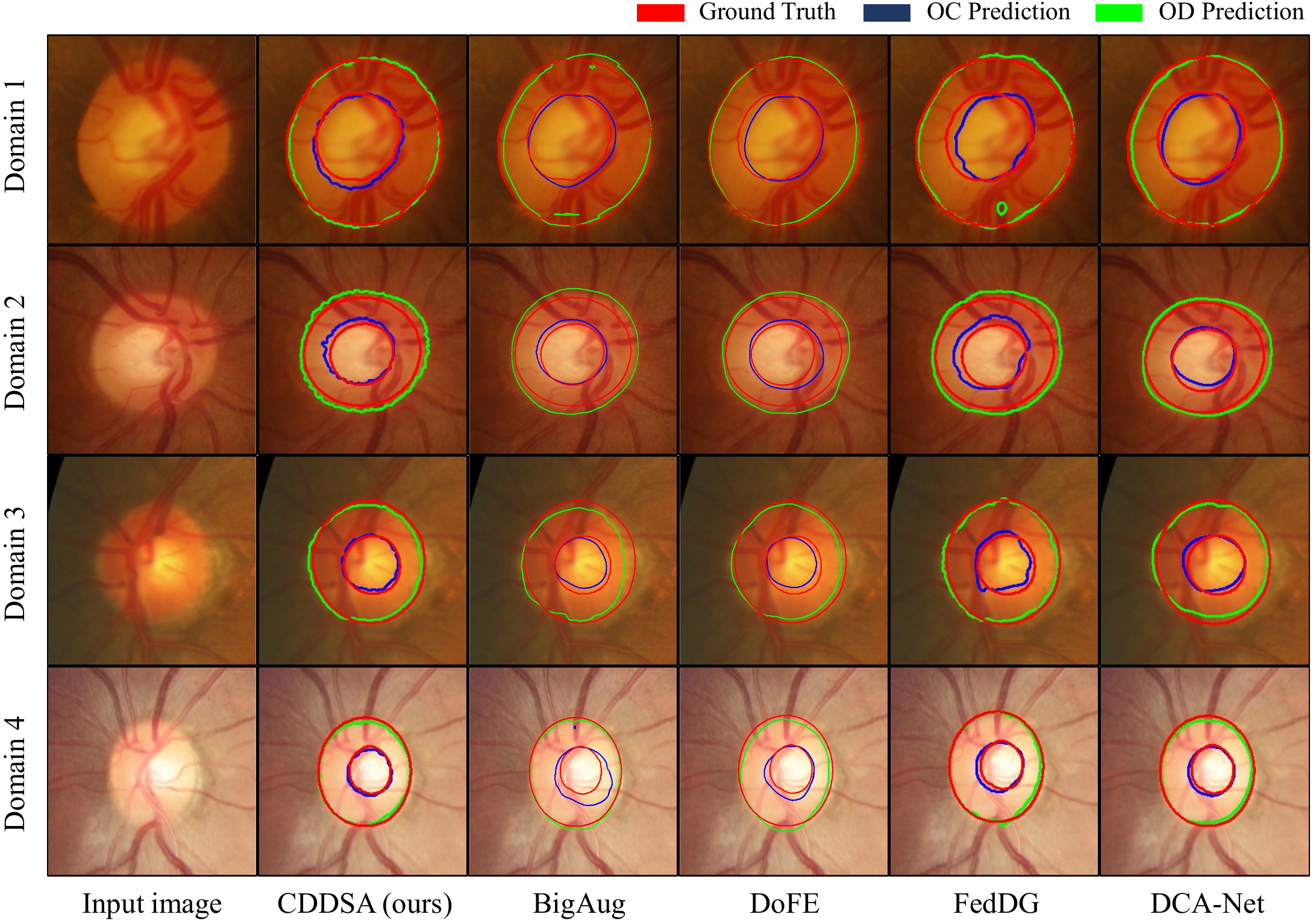}
    \caption{Visual comparison between our proposed CDDSA and BigAug~\citep{zhang2020generalizing}, DoFE~\citep{wang2020dofe}, FedDG~\citep{liu2021feddg} and DCA-Net~\citep{gu2021domain} on multi-domain fundus image segmentation.} 
    \label{fig2:fundus_results}
\end{figure*}
\begin{table*}
    % \small
    \centering
    \caption{Comparison between different activation functions used by the output of $E_{ana}$ for multi-domain OC$/$OD segmentation. Gumbel-H and Gumbel-S are two variants of gumbel softmax that return discrete one-hot values and soft continuous  values, respectively.}\label{tab3:fundus_ablation}
    \scalebox{0.8}{\begin{tabular}{l|l|c|c|c|c|c|c|c|c|c|c}
    \hline
    \hline
    \multirow{2}{0.3in}{Metric} &
    \multirow{2}{0.3in}{Activation} & \multicolumn{2}{c}{Domain 1} & \multicolumn{2}{|c|}{Domain 2} & \multicolumn{2}{|c|}{Domain 3} & \multicolumn{2}{|c|}{Domain 4} & \multicolumn{2}{|c}{Avg} \\ \cline{3-12}
    &  & cup & disc & cup & disc & cup & disc & cup & disc & cup & disc \\ \hline
    \multirow{4}{0.3in}{Dice} & Gumbel-H & 82.29$\pm$11.97 & 96.46$\pm$2.11 & 78.67$\pm$18.79 & 86.21$\pm$5.06 & 85.40$\pm$8.14 & \textbf{93.36$\pm$3.19} & \textbf{87.89$\pm$7.67} & 93.22$\pm$3.16 & 83.56 & 92.31 \\
    & Gumbel-S &
    82.88$\pm$11.28 & 96.71$\pm$1.65 & 80.22$\pm$13.76 & 89.11$\pm$3.52 &  83.20$\pm$8.90 & 92.59$\pm$4.91 & 86.41$\pm$9.88 & 94.39$\pm$3.53 & 83.18 & 93.20 \\
    & softmax & 85.22$\pm$10.46 & \textbf{96.94$\pm$1.26} & 80.72$\pm$16.09 & 88.42$\pm$12.16 &  85.22$\pm$7.43 & 93.13$\pm$3.75 & 85.34$\pm$10.07 & 93.23$\pm$3.78 & 84.13 & 92.93 \\
    & tanh & \textbf{85.75$\pm$12.31} & 96.79$\pm$1.53 & \textbf{81.04$\pm$13.63} & \textbf{89.71$\pm$3.60} &  \textbf{86.94$\pm$7.94} & 93.25$\pm$3.55 & 86.86$\pm$8.97 & \textbf{94.44$\pm$3.96} & \textbf{85.15} & \textbf{93.55} \\ \hline
    
    \multirow{4}{0.3in}{ASSD} & Gumbel-H & 18.31$\pm$8.49 & 6.65$\pm$4.11 & 14.60$\pm$17.21 & 18.13$\pm$16.60 &  9.92$\pm$5.45 & \textbf{9.10$\pm$3.66} & \textbf{6.67$\pm$3.96} & 6.97$\pm$3.02 & 12.38 & 10.21 \\
    & Gumbel-S & 17.22$\pm$6.91 & \textbf{6.33$\pm$2.83} & \textbf{12.30$\pm$6.02} & \textbf{12.98$\pm$5.03} &  10.94$\pm$5.57 & 9.78$\pm$5.85 & 7.52$\pm$5.42 & 6.89$\pm$4.77 & 12.00 & 9.00 \\
    & softmax & \textbf{14.40$\pm$6.36} & 6.67$\pm$2.53 & 12.61$\pm$12.71 & 13.28$\pm$12.93 &  10.01$\pm$5.14 & 9.90$\pm$5.09 & 8.28$\pm$5.90 & 7.67$\pm$4.73 & 11.33 & 9.38 \\
    & tanh & 14.65$\pm$8.39 & 6.54$\pm$3.74 & 12.91$\pm$10.79 & 13.06$\pm$8.60 &  \textbf{9.38$\pm$5.40} & 9.32$\pm$4.11 & 7.28$\pm$5.85 & \textbf{6.87$\pm$5.03} & \textbf{11.06} & \textbf{8.95} \\
    \hline
    \hline
    \end{tabular}}
\end{table*}

\subsection{Fundus Image Segmentation}
\subsubsection{Comparison with State-of-the-art DG Methods}
\label{sec:fundus_sota}
For domain generalization study, we conducted leave-one-domain-out cross validation on the multi-domain fundus image dataset. We first considered all the available training domains as a single dataset (i.e., ignoring the domain shift in training set) and trained a U-Net~\citep{ronneberger2015u} using a standard Dice loss, and directly applied it to the unseen domain, which is referred to as `\textbf{Inter-domain}' and serves as a lower bound of the experiment. Then, for each domain, we trained and tested the U-Net~\citep{ronneberger2015u} with the training and testing sets respectively,  i.e., no unseen domain involved, which serves as the upper bound for DG and is referred as `\textbf{Intra-domain}'. For DG methods, we compared our proposed CDDSA with four representative state-of-the-art approaches: BigAug~\citep{li2018domain} based on  data augmentation, DoFE~\citep{wang2020dofe} based on domain-oriented feature embedding,  DCA-Net~\citep{gu2021domain} based on  domain composition and attention, and FedDG~\citep{liu2021feddg} that is a federated learning-based domain generalization method.

Table~\ref{tab1:fundus_dice} and Table~\ref{tab2:fundus_asd} show the quantitative evaluation results of OC$/$OD segmentation in terms of Dice and ASSD, respectively. Intra-domain achieved the highest performance among the compared methods, with an average Dice of 85.33$\%$ and 95.55$\%$ for the OC and OD across the four domains. %, and their corresponding average ASSD was 10.91 and 7.04 pixels. 
In contrast, the average Dice achieved by Inter-domain was only 79.61$\%$ and 90.32$\%$ % and 14.47 and 11.58 pixels 
in OC and OD segmentation, respectively, showing the performance gap caused by domain shift. BigAug~\citep{li2018domain} obtained a slight improvement from Inter-domain, suggesting that aimlessly conducting data augmentation in the image domain has a limited performance. %In contrast, DoFE and FedDG performed better than BigAug, but they only implicitly utilize these domain discriminative representations and do not introduce domain constraints to obtain more thorough domain-specific representations. Introducing domain-invariant learning and domain divergence constraint, 
Among the compared existing methods, DCA-Net~\citep{gu2021domain} achieved the highest performance, with an average Dice of 83.60$\%$ and 92.64$\%$ for OC and OD, respectively. %However, without the help of domain style contrastive learning, DCA-Net just encourages the extracted representations from different domains to be as divergent as possible while not controlling a proper distance in a high dimensional space. 
In contrast, our proposed CDDSA outperformed the existing methods, with an average Dice of  85.15$\%$ and 93.55$\%$ for OC and OD, respectively. 
The average ASSD obtained by our method was  11.06 and 8.95 pixels for OC and OD, respectively, which also outperformed the compared methods, as shown in Table~\ref{tab2:fundus_asd}. 
Fig.~\ref{fig2:fundus_results} shows a visual comparison between our proposed CDDSA and BigAug, DoFE, FedDG and DCA-Net for images from the four  testing domains, respectively. It shows that the segmentation results obtained by our proposed CDDSA had  boundaries that are closer to the ground truth, while the other DG methods have more over- and under-segmented regions than ours.
\begin{figure*}
    \centering
    \includegraphics[width=0.82\textwidth]{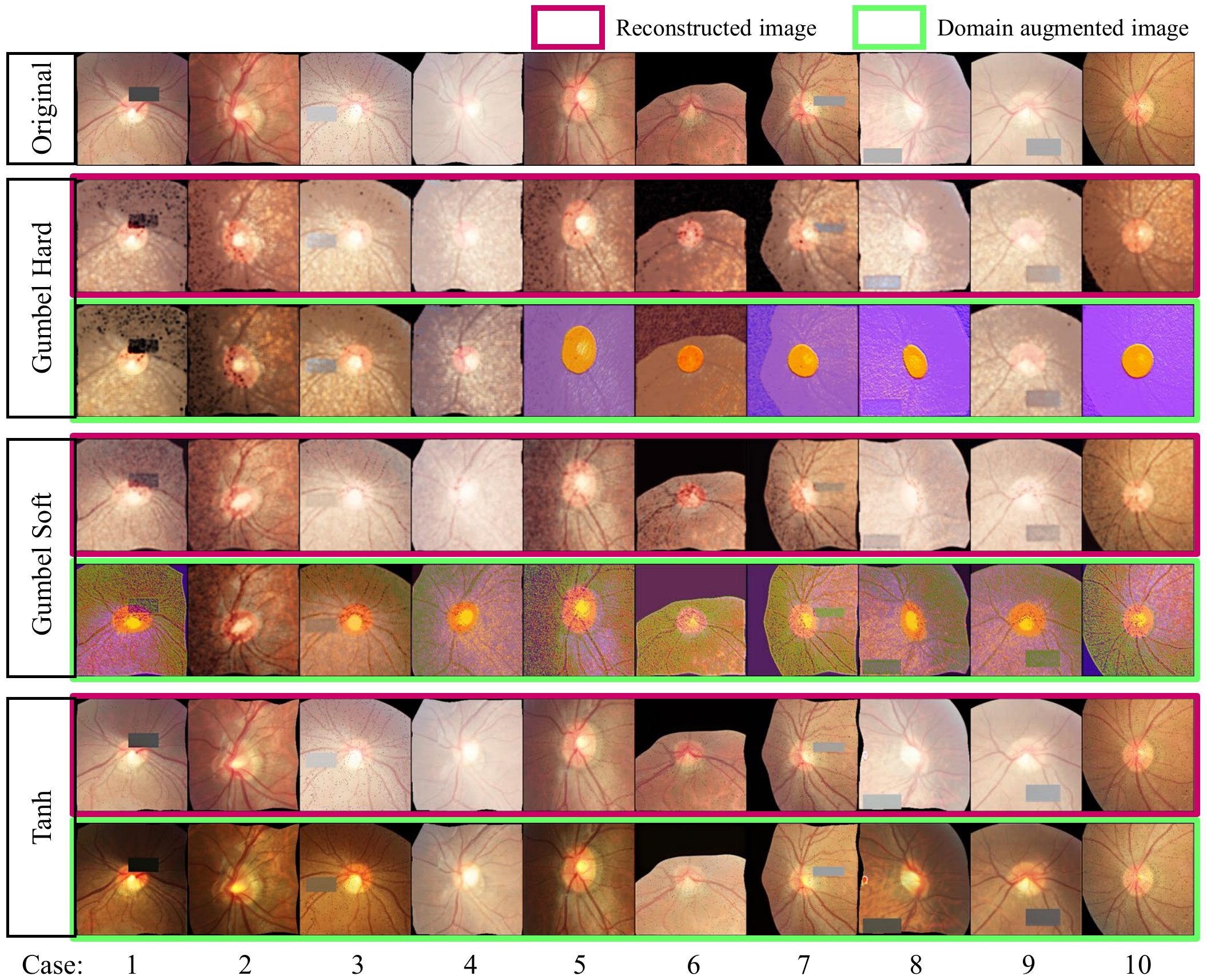}
    \caption{Visual comparison of 
    reconstructed  and augmented fundus images with different activation functions  at the end of anatomy Encoder. The original images are from domain 4. %Here, we make comparison on disentangle baseline SDNet~\citep{chartsias2019disentangled} recommending gumbel softmax and our recommending tanh. 
    For each method, the first row (red rectangles) shows images reconstructed from the disentangled anatomical representation and style code, and the second row (green rectangles) shows style-augmented images that are generated based on the anatomical representation from the original images and changed style codes.
    }
    \label{fig3:fundus_reconstruct_quality}
\end{figure*}

\subsubsection{Ablation Studies}
\textbf{Effectiveness of Domain Style Contrastive Learning and Style Augmentation:} 
We conducted ablation studies to evaluate the effectiveness of the components of our CDDSA framework, where the baseline was only training $E_{ana}$, $E_{sty}$, $D_{rec}$ and $S$ with basic loss functions of $\mathcal{L}_{seg}$, $\mathcal{L}_{kl}$ and $\mathcal{L}_{rec}$, following SDNet~\cite{chartsias2019disentangled}. We use +$\mathcal{L}_{dsct}$ and +$\mathcal{L}_{saac}$ to denote adding the domain style contrastive learning and domain style augmentation with anatomical consistency to the baseline, respectively. +$\mathcal{L}_{dsct}$ +$\mathcal{L}_{saac}$ means our proposed CDDSA. 

%We proved the effectiveness of our proposed domain style contrastive learning and style augmentation strategy. The baseline is actually modified SDNet~\citep{chartsias2019disentangled} for feature disentanglement, in which we only input original image into $E_{sty}$ to obtain the anatomical representation $f_{a}^{d}$. $\mathcal{L}_{seg}$, $\mathcal{L}_{kl}$ and $\mathcal{L}_{rec}$ are used in the baseline. +$\mathcal{L}_{dsct}$ means baseline introduced with domain style contrastive learning; +$\mathcal{L}_{saac}$ means baseline with style augmentation and anatomical consistency; and +$\mathcal{L}_{dsct}$+$\mathcal{L}_{saac}$ means baseline with both domain style contrastive learning and style augmentation with anatomical consistency.
Quantitative evaluation results in terms of Dice and ASSD of these variants are shown in the last section of Table~\ref{tab1:fundus_dice} and Table~\ref{tab2:fundus_asd}, respectively.
The baseline's average Dice across OC and OD was 87.42$\%$, and combining the baseline with $\mathcal{L}_{dsct}$ improved it to 88.81$\%$, indicating that encouraging the network to obtain more discriminative style codes leads to better generalization performance. Baseline + $\mathcal{L}_{saac}$ also improved the two classes' average Dice to 88.48\%, and our method using $\mathcal{L}_{dsct}$ and  $\mathcal{L}_{saac}$ further improved the average Dice to 89.35\% (85.15\% for OC and 93.55\% for OD), showing the extra improvement brought by the proposed style augmentation. 

To additionally evaluate the effectiveness of our proposed random linear combination for generating new style code during style augmentation, we compared it with an alternative method that randomly samples the style code from a  Gaussian distribution, which is denoted as CDDSA$^\diamond$ in Table~\ref{tab1:fundus_dice} and Table~\ref{tab2:fundus_asd}. The results showed CDDSA$^\diamond$ performed slightly worse than CDDSA, but outperformed most existing DG methods, proving that our proposed random linear combination was better for style augmentation than randomly sampling the style code from a Gaussian distribution.
\begin{table*}
    % \small
    \centering
    \caption{Quantitative comparison of different DG methods on the multi-domain NPC-MRI image dataset for GTVnx segmentation. % +$\mathcal{L}_{dsct}$ means baseline network introduces style contrastive learning. +$\mathcal{L}_{saac}$ means baseline network introduces style augmentation strategy. +$\mathcal{L}_{dsct}$+$\mathcal{L}_{saac}$ means baseline network introduces both style contrastive learning and style augmentation strategy that is our final proposed CDDSA model. CDDSA$\diamond$ means  the new style code was randomly sampled from a Gaussian distribution rather than using random linear combination style codes in the source domain.
    }
    \label{tab4:npc_dice_assd}
    \scalebox{0.75}{\begin{tabular}{l|c|c|c|c|c|c|c|c|c|c}
    \hline
    \hline
    \multirow{2}{0.9in}{Methods} & \multicolumn{2}{c}{Domain 1} & \multicolumn{2}{|c|}{Domain 2} & \multicolumn{2}{|c|}{Domain 3} & \multicolumn{2}{|c|}{Domain 4} &
    \multicolumn{2}{|c}{Avg} \\ \cline{2-11}
    & Dice (\%) & ASSD (pix) & Dice (\%)  & ASSD (pix) & Dice (\%)  & ASSD (pix) & Dice (\%)  & ASSD (pix) & Dice  & ASSD\\ \hline
    Lower bound (Inter-domain) & 
    65.44$\pm$13.35 & 2.37$\pm$1.52 &
    76.65$\pm$7.75 & 1.70$\pm$1.06 & 
    81.17$\pm$7.01 & 1.61$\pm$0.77 & 
    61.62$\pm$13.66 & 3.38$\pm$1.47 & 
    71.22 & 2.27 \\
    
    Upper bound (Intra-domain) & 
    79.19$\pm$6.35 & 1.13$\pm$0.54 & 
    82.89$\pm$7.54 & 1.58$\pm$1.75 & 
    86.30$\pm$3.25 & 1.26$\pm$0.48 & 
    79.46$\pm$7.40 & 2.09$\pm$1.48 & 
    81.96 & 1.52 \\ \hline
    
    BigAug \citep{zhang2020generalizing} & 
    75.63$\pm$5.97 & 1.67$\pm$1.01 & 
    78.51$\pm$7.89 & 1.65$\pm$0.99 &
    82.30$\pm$5.05 & 1.82$\pm$0.66 &
    63.88$\pm$12.32 & 4.05$\pm$1.71 & 
    75.08 & 2.30 \\
    
    DoFE \citep{wang2020dofe} & 
    \textbf{78.44$\pm$6.97} & \textbf{1.27$\pm$0.99} & 
    75.00$\pm$4.95 & 1.80$\pm$0.85 & 
    79.66$\pm$5.77 & 1.66$\pm$0.68 & 
    64.71$\pm$15.06 & 2.57$\pm$1.91 & 
    74.45 & 1.83 \\ 
    
    FedDG \citep{liu2021feddg} & 
    65.07$\pm$11.59 & 2.58$\pm$1.63 & 
    78.90$\pm$6.11 & 1.67$\pm$1.09 & 
    81.57$\pm$6.05 & 1.79$\pm$0.79 & 
    \textbf{72.38$\pm$12.13} & 3.91$\pm$2.64 & 
    74.48 & 2.49 \\
    
    DCA-Net \citep{gu2021domain} & 
    77.27$\pm$6.66 & 1.27$\pm$0.99 & 
    77.14$\pm$7.53 & 1.80$\pm$0.85 & 
    81.63$\pm$6.20 & 1.66$\pm$0.68 & 
    69.32$\pm$10.08 & 2.57$\pm$1.91 & 
    76.34 & 1.83 \\ \hline
    
    Baseline & 
    76.63$\pm$5.74 & 1.53$\pm$1.23 & 
    77.93$\pm$5.99 & \textbf{1.41$\pm$0.72} &
    82.77$\pm$4.65 & 1.55$\pm$0.55 & 
    62.71$\pm$11.63 & 3.07$\pm$1.91 &
    75.01 & 1.89 \\
    
    +$\mathcal{L}_{dsct}$ & 
    77.02$\pm$6.09 & 1.56$\pm$1.09 &
    77.11$\pm$7.15 & 1.71$\pm$0.89 &
    83.02$\pm$3.62 & 1.45$\pm$0.41 & 
    63.88$\pm$12.74 & 2.81$\pm$1.66 & 
    75.26 & 1.88 \\
   
    +$\mathcal{L}_{saac}$ & 
    77.74$\pm$5.67 & 1.35$\pm$0.82 &
    76.33$\pm$8.37 & 1.66$\pm$0.83 &
    83.23$\pm$5.38 & \textbf{1.44$\pm$0.56} & 
    68.43$\pm$12.92 & 3.45$\pm$1.99 &
    76.43 & 1.98 \\
    
    +$\mathcal{L}_{dsct}$+$\mathcal{L}_{saac}$ (CDDSA$\diamond$) & 77.77$\pm$6.65 & 1.36$\pm$0.97 & 
    78.00$\pm$7.74 & 1.52$\pm$0.86 & 
    83.10$\pm$5.18 & 1.60$\pm$0.63 & 
    66.39$\pm$10.88 & 3.50$\pm$1.62 & 
    76.32 & 2.00 \\
    
    +$\mathcal{L}_{dsct}$+$\mathcal{L}_{saac}$ (CDDSA) & 78.34$\pm$5.14 & 1.37$\pm$0.82 & 
    \textbf{79.16$\pm$6.68} & 1.61$\pm$1.19 & 
    \textbf{83.53$\pm$4.55} & 1.48$\pm$0.54 & 
    69.53$\pm$10.28 & \textbf{2.46$\pm$1.50} & 
    \textbf{77.64} & \textbf{1.73} \\
    \hline
    \hline
    \end{tabular}}
\end{table*}
\begin{figure*}
    \centering
    \includegraphics[width=1.0\textwidth]{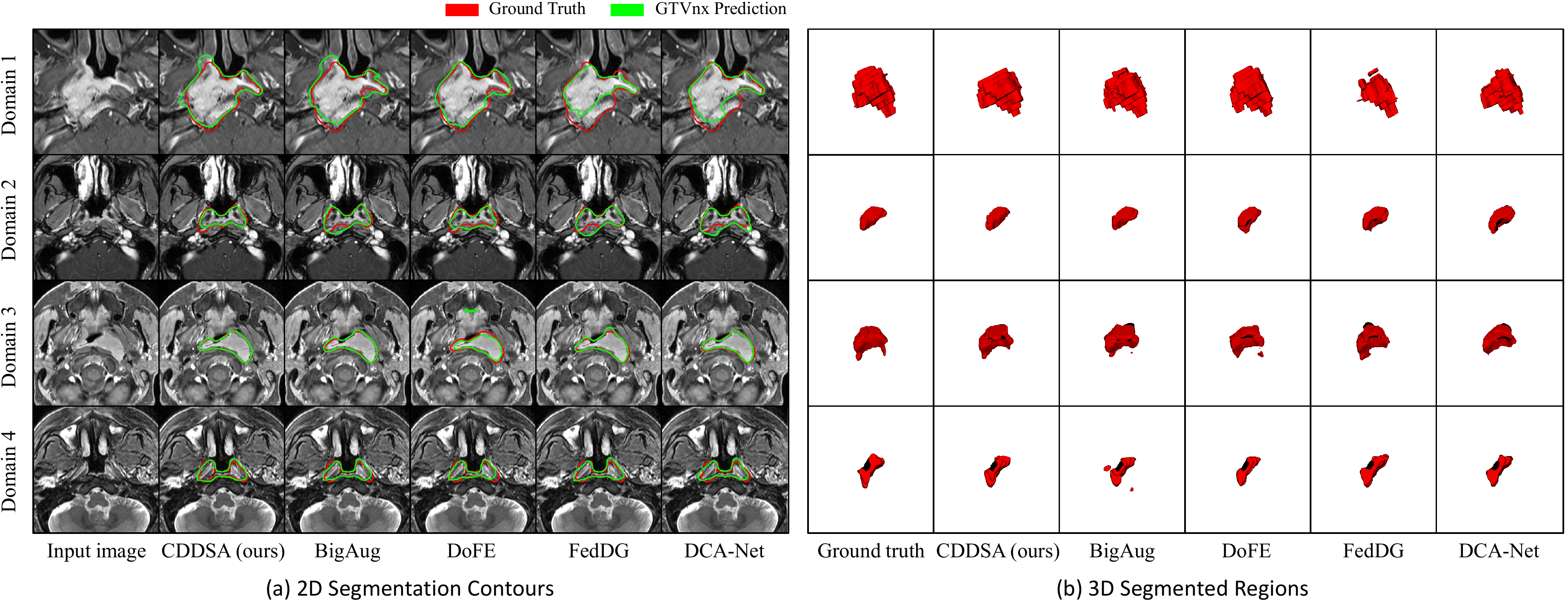}
    \caption{Visual comparison between different DG methods for multi-domain NPC GTVnx segmentation.} 
    \label{fig4:npc_results}
\end{figure*}

\textbf{Reconstruction and Style Augmentation Quality:}
\label{sec:fundus_ablation}
Since the decomposed anatomical representation $f_{a}$ serves as the input of the segmentor $S$ and the reconstruction decoder $D_{rec}$, the quality of the anatomical representation has an impact on the performance of $S$ and  $D_{rec}$. To explore the influence of different 
formats of the anatomical representation on reconstruction and segmentation quality, we compared four activation functions at the end of $E_{ana}$: 1) the gumbel softmax~\citep{JangGP17} returning discrete one-hot values, which is referred to as Gumbel-H; 2) the gumbel softmax returning continuous soft values, which is referred to as Gumbel-S; 3) Softmax and 4)  Tanh. 
Quantitative comparison of these actuation functions in the fundus image segmentation task is shown in Table~\ref{tab3:fundus_ablation}. We found that Gumbel-S obtained a better segmentation performance than Gumbel-H (83.56$\%$ and 92.31$\%$ vs 83.18$\%$ and 93.20$\%$ of average Dice score). Softmax and tanh further improved the model's performance. Notably, Tanh  achieved the highest average Dice score of 85.15$\%$ in OC and 93.55$\%$ in OD and the lowest average ASSD (11.06 pixels in OC and 8.95 pixels in OD) among the compared activation functions. The results show that using continuous soft values for anatomical representations led to better segmentation performance, as  soft representations are more informative compared with binary representations~\citep{chartsias2020disentangle}.

To further investigate how the activation function used by $E_{ana}$ affects the reconstructed  images and style augmentation, %To comprehensively support that utilizing consecutive soft activation functions is beneficial for style augmentation reconstruction and domain generalization segmentation, 
we compared the original training image with the image reconstructed  from disentangled $f_{a}$ and $f_{s}$ and the style-augmented image in Fig~\ref{fig3:fundus_reconstruct_quality}, where we show the differences between Gumbel-H, Gumbel-S and Tanh. %, where all the original images are from domain 4. %visualized the reconstructed images from disentangled $f_{a}$ and $f_{s}$ (the first row of each method) as well as our style-augmented images (the second row of each method) in domain 4 as shown in Fig.~\ref{fig3:fundus_reconstruct_quality}. 
%To compare the adaptation of our introduced tanh activation functions with the previously utilized gumbel softmax in disentangle baseline for domain generalization segmentation, we mainly showed the visual comparison of reconstructions and style augmentations. 
%We mainly presented the visual comparison of our proposed tanh activation and previous recommended gumbel softmax activation.
First, for image reconstruction (even rows), it can be observed that using Gumbel-H only reconstructed coarse-grained images and only roughly retrained the overall content without detailed structures. Gumbel-S achieved a better quality with more details than Gumbel-H. However, the reconstructed images have some noticeable artefacts compared with the original images. In contrast, Tanh obtained a much higher quality than Gumbel-H, and the reconstructed images were closer to the original inputs in terms of both anatomical structures and styles. Second, for style augmentation (odd rows), Gumbel-H can not keep the same anatomical structure after changing the style, and led to unrealistic images in the augmented domain, especially in cases 5, 7, 8 and 10 in the third row of Fig.~\ref{fig3:fundus_reconstruct_quality}. 
%Notably, as the aspect of evaluating the efficiency and reconstructing quality of our proposed style augmentation, all methods have possessed the ability to generate a new style image based on the decomposed anatomical representations $f_{a}$ from the source domain when introducing our proposed style augmentation strategy, especially the noticeable effect presented in case 3, 4, 8 and 9 of Fig.~\ref{fig3:fundus_reconstruct_quality}. 
%In comparison, relying on discretized features as $f_{a}$ will reconstruct bad-case samples that are distorted, as shown in case 5, 7, 8 and 10 of Gumbel Hard in Fig.~\ref{fig3:fundus_reconstruct_quality}. Although utilizing gumbel softmax activation with returning soft features as $f_{a}$ avoids this issue, the reconstructed images lose redundant content in details.
Gumbel-S has a better ability to remain the anatomical structures, but the augmented images have a lot of artefacts with unrealistic appearance. In contrast, Tanh achieved very high quality in the style-augmented images  with realistic appearances. They have quite different styles with shared anatomical structures compared with the original images, as shown in the last row of Fig.~\ref{fig3:fundus_reconstruct_quality}. %This demonstrates that our style-augmented images greatly retain the anatomy structure from the source domain while at the same time successfully transferring the style distribution to an unseen field. 
The results show the advantage of our proposed style augmentation strategy, which can successfully generate new samples in an unknown domain with anatomical structures unchanged, which is beneficial for enhancing the model's generalizability. % without introducing vanilla generative methods filling in complicated training procedures, e.g., Generative Adversarial Network (GAN).

\subsection{NPC GTVnx Image segmentation}
\subsubsection{Comparison with State-of-the-art DG Methods}
For the multi-domain GTVnx segmentation task, we employed the same set of methods as in Section~\ref{sec:fundus_sota} for comparison, and the quantitave evaluation results are shown in Table~\ref{tab4:npc_dice_assd}. %Since the images were segmented slice-by-slice with 2D networks, we stacked the segmentation in the slices to reconstruct a 3D result for evaluation.%Here, we employed the Dice score and Average Symmetric Surface Distance (ASSD) for quantitative evaluation.
%Comprehensive evaluated results in GTVnx generalizable segmentation are shown in Table.~\ref{tab4:npc_dice_assd}. 
First, Intra-domain (upper bound) achieved the highest performance with average Dice of 81.96$\%$ and average ASSD of 1.52 mm across the four domains. In contrast, Inter-domain (lower bound) only obtained an average Dice of 71.22$\%$ and ASSD of 2.27 mm. The performance gap between them was over 10$\%$ in average Dice, indicating the large  shift among the different domains. All the four existing DG methods achieved great improvements compared with the Inter-domain that does not consider the differences across domains. DoFE~\citep{wang2020dofe} and FedDG~\citep{liu2021feddg} had a similar segmentation performance, with average Dice score of 74.45$\%$ and 74.48$\%$, respectively. BigAug~\citep{zhang2020generalizing} achieved an average Dice of 75.08\%, indicating that some augmentation strategies are beneficial for GTVnx segmentation in cross-modality MRI images. DCA-Net~\citep{gu2021domain}    obtained an average Dice  of 76.34$\%$, which outperformed the other three existing DG methods. In contrast, our proposed CDDSA based on domain-invariant feature learning obtained higher generalizability, achieving an average Dice of 77.64$\%$ and ASSD of 1.73~mm, which outperformed the state-of-the-art DG methods. %The average Dice score of our proposed CDDSA is only a 4.32$\%$ performance gap compared with the upper-bound method 'Intra-domain'. 

Fig.~\ref{fig4:npc_results} provides a visual comparison between our proposed CDDSA  and the four state-of-the-art DG methods on the multi-domain NPC GTVnx segmentation dataset. %, where (a) and (b) show the results in 2D and 3D, respectively. 
Fig.~\ref{fig4:npc_results}(a) shows that the 2D segmentation boundaries of our CDDSA are closer to the ground truth than those of the other methods.  The 3D visualization in Fig.~\ref{fig4:npc_results} (b) shows that our CDDSA achieved high-quality segmentation results, while the other DG methods have more noises in the  results.
\begin{table*}
    % \small
    \centering
    \caption{Comparison between different activation functions used by the output of $E_{ana}$ for multi-domain NPC GTVnx segmentation.}\label{tab5:npc_ablation}
    \scalebox{0.8}{\begin{tabular}{l|c|c|c|c|c|c|c|c|c|c}
    \hline
    \hline
    Activation & \multicolumn{2}{c}{Domain 1} & \multicolumn{2}{|c|}{Domain 2} & \multicolumn{2}{|c|}{Domain 3} & \multicolumn{2}{|c|}{Domain 4} & \multicolumn{2}{|c}{Avg} \\ \cline{2-11}
     & Dice (\%) & ASSD (pix) & Dice (\%) & ASSD  (pix) & Dice (\%) & ASSD (pix) & Dice (\%) & ASSD (pix) & Dice (\%) & ASSD (pix) \\ \hline
    Gumbel-H & 
    75.60$\pm$6.20 & 1.61$\pm$1.15 & 
    72.69$\pm$9.42 & 1.60$\pm$0.76 & 
    82.81$\pm$7.67 & 1.62$\pm$0.58 &
    64.73$\pm$13.53 & 2.90$\pm$1.88 &
    73.96 & 1.93 \\
    
    Gumbel-S &
    76.54$\pm$6.39 & 1.48$\pm$1.10 & 
    78.40$\pm$6.48 & 1.57$\pm$0.88 & 
    83.23$\pm$4.54 & \textbf{1.47$\pm$0.52} & 65.24$\pm$11.78 & 2.73$\pm$1.43 & 
    75.88 & 1.81  \\
    
    softmax & 
    77.46$\pm$5.40 & 1.57$\pm$1.01 & 
    \textbf{79.87$\pm$4.70} & \textbf{1.46$\pm$0.74} & 
    82.85$\pm$4.89 & 1.89$\pm$0.91 & 62.59$\pm$12.48 & 3.08$\pm$1.59 & 
    75.69 & 2.00  \\
    
    tanh & 
    \textbf{78.34$\pm$5.14} & \textbf{1.37$\pm$0.82} & 
    79.16$\pm$6.68 & 1.61$\pm$1.19 & 
    \textbf{83.53$\pm$4.55} & 1.48$\pm$0.54 & \textbf{69.53$\pm$10.28} & \textbf{2.46$\pm$1.50} & 
    \textbf{77.64} & \textbf{1.73} \\
    \hline
    \hline
    \end{tabular}}
\end{table*}
\begin{figure*}
    \centering
    \includegraphics[width=0.96\textwidth]{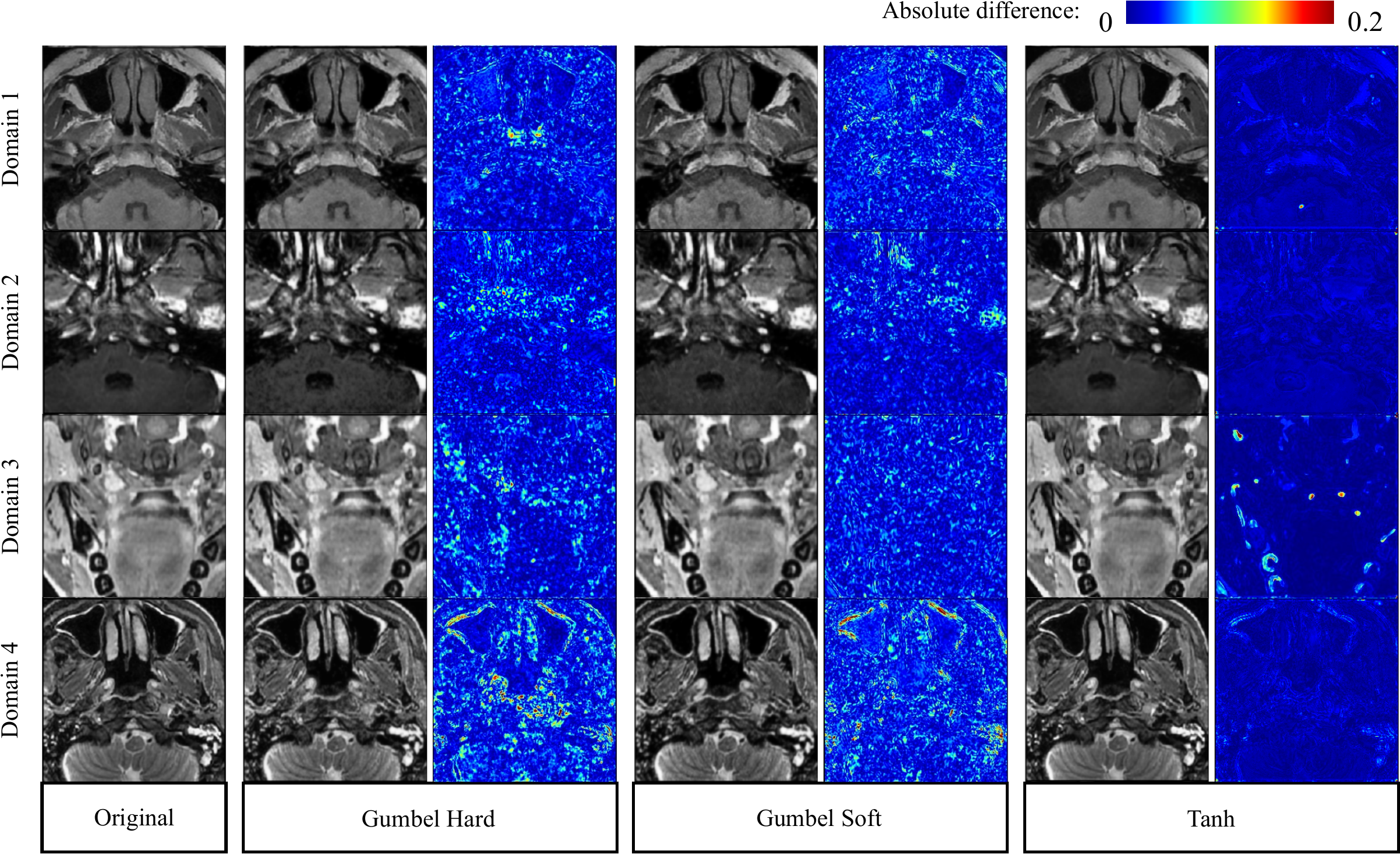}
    \caption{Comparison of reconstructed images with different activation functions at the end of $E_{ana}$. For each method, the first column  shows the reconstructed images based on the disentangled anatomical representation and style code, and the second column shows the absolute difference between the reconstructed and original images.}
    \label{fig5:npc_reconstruct_quality}
\end{figure*}
\begin{figure*}
    \centering
    \includegraphics[width=1.0\textwidth]{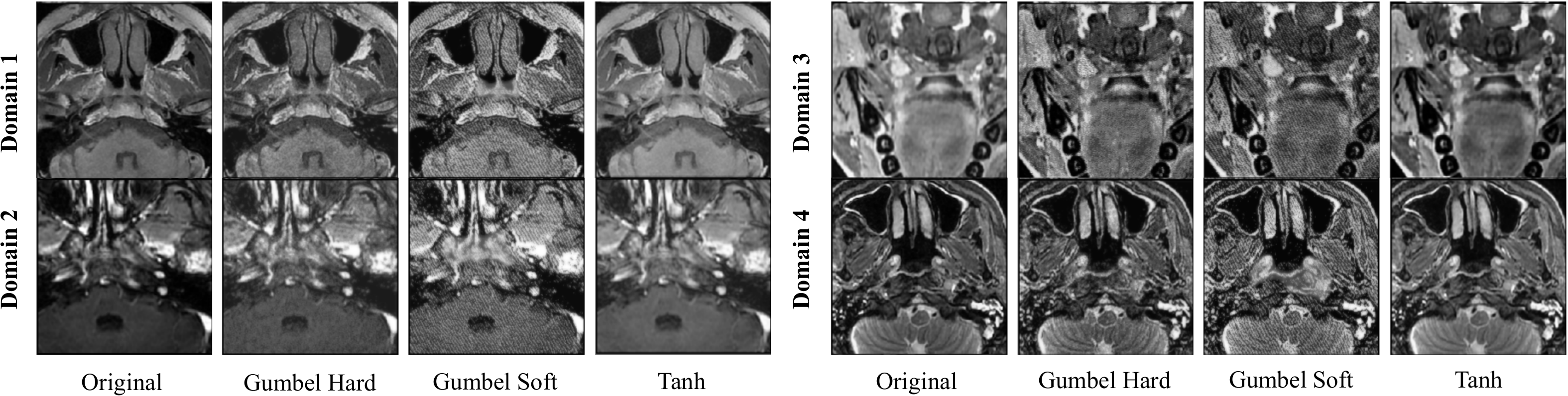}
    \caption{Visual comparison of style-augmented images with different activation functions at the end of $E_{ana}$. %We make comparison on disentangle baseline recommending gumbel softmax and our recommending tanh.
    }
    \label{fig6:npc_augmented_img}
\end{figure*}

\subsubsection{Ablation Studies}
\textbf{Effectiveness of Domain Style Contrastive Learning and Style Augmentation:} Similar with multi-site fundus image segmentation. We also proved the effectiveness of our proposed domain style contrastive learning and style augmentation strategy in multi-site NPC GTVnx segmentation. Quantitative results are shown in Table~\ref{tab4:npc_dice_assd}.
The baseline, i.e., re-implementation of SDNet~\citep{chartsias2019disentangled} based on our network structures, obtained an average Dice of 75.01$\%$, and combining it with our domain style contrastive learning $\mathcal{L}_{dsct}$ improved it to 75.26$\%$.  Combining it with our domain augmentation method $\mathcal{L}_{saac}$ achieved an average Dice of 76.43\%. In contrast, our proposed method that uses  $\mathcal{L}_{dsct}$  and $\mathcal{L}_{saac}$ simultaneously improved the average Dice to 77.64$\%$, which is the highest among the compared variants and significantly better than the baseline ($p$-value $<$ 0.05). %, indicating that our proposed CDDSA builds the comprehensive model generalization when testing on unseen target projects. Similarly,
Table~\ref{tab4:npc_dice_assd} also shows that CDDSA performed better than CDDSA$\diamond$ (77.64\% vs 76.32\%) in terms of Dice, indicating that our style augmentation based on random linear combination of the style codes was better  than directly sampling style codes from a Gaussian distribution for style augmentation.

\textbf{Reconstruction and Style Augmentation Qualities:}
Similar to Section~\ref{sec:fundus_ablation}, we compared four different activation functions at the end of $E_{ana}$ to represent $f_{a}$ on the NPC-MRI dataset. The corresponding  NPC-MRI GTVnx segmentation results are shown in Table~\ref{tab5:npc_ablation}. It can be observed that Gumbel-S had a higher performance than Gumbel-H (75.88$\%$ vs 73.96$\%$ in terms of average Dice). Using Tanh further improved the average Dice to 77.64\%, which was significantly better than the other activations. %Notably, the tanh activation function achieves the most remarkable average Dice score of 77.64\% andF the smallest average ASSD (1.73 mm) among all activation functions. Quantitative experimental results prove that remaining consecutive soft features as the output anatomical representations obtains more remarkable segmentation performance, especially for tanh.

Fig.~\ref{fig5:npc_reconstruct_quality} shows a visual comparison of these activation functions in reconstructing the original image after disentanglement. It can be observed that when Gumbel-H is used, the reconstructed images have a large difference from the original images. Gumbel-S has a lower reconstruction error than Gumbel-H. However, it is inferior to our method using Tanh, showing that Tanh is more suitable to obtaining anatomical representation in disentanglement for high-fidelity reconstruction. 
%to intuitively compare the reconstruction quality of using different activation functions in $E_{ana}$, we visualized the reconstructed images $\hat{x}^d$ from disentangled $f_{a}$ and $f_{s}$ (the first column of each method) as well as their corresponding heat maps of absolute difference computed between reconstructed images and original images (the second column of each method) in 4 domains. Similar to Fig.~\ref{fig3:fundus_reconstruct_quality}, we presented a visual comparison of disentangle baseline recommending gumbel softmax and our recommending tanh activation functions in NPC-MRI GTVnx segmentation as shown in Fig.~\ref{fig5:npc_reconstruct_quality}. In disentangle reconstruction, although the previously used gumbel softmax, either returning discretized or consecutive features, successfully reconstructs whole images, it roughly retrained the overall content without detailed structures. The corresponding absolute difference between the reconstructed and original images is significantly visible in the heat map manner, where plenty of areas exist distinguishable reconstructive differences. In contrast, our used $f_{a}$ output by tanh activation reconstructed the most fine-gained images and their heat map are more clear than gumbel softmax activations.
%However, in terms of imaging quality, they recover the details of the image reluctantly. In contrast, using the tanh activation function obtains the fine gained images with the most detailed structure reduction ability among them.

In addition, Fig.~\ref{fig6:npc_augmented_img} shows a visual comparison of  style-augmented images when different activation functions are used at the end of $E_{ana}$. We found that all the methods can generate new-style images based on the augmented domain style code $\tilde{f}_{s}$ and the anatomical representation $f_{a}$ of the input. However, Gumbel-H and Gumbel-S led to obvious artifacts in the augmented images. In contrast, our method can change the style of an input image while better retaining the anatomical structures. 
%Concretely, relying on gumbel softmax activated features as $f_{a}$ will reconstruct slightly coarse-grained samples compared with the original images. This issue is extremely distinct in reconstruction of utilizing gumbel softmax with returning soft feature. Notably, our tanh activation achieves inspiring reconstructive quality of domain augmented images, as shown in the last column of each domain in Fig.~\ref{fig6:npc_augmented_img}. Encouragingly, the restored augmented samples greatly retain the anatomical structures from the source domain while successfully transferring the style distribution to the unseen field.

\section{Conclusion}
In this paper, we present a Contrastive Domain Disentanglement and Style Augmentation  (CDDSA) framework to tackle the domain generalization problem in medical image segmentation. %for the purpose of decomposing out the domain-invariant anatomical features across domain. 
We introduce a GAN-free efficient disentangle method to decompose medical images from multiple domains into a domain-invariant anatomical representation and a domain-specific style code, where a segmentor works on the anatomical representation to achieve generalizability.  To improve the disentanglement and segmentation performance, we use a soft representation for the anatomical representation based on Tanh, and propose domain  style contrastive learning to minimize the similarity of style codes in different domains. Based on the disentanglement, we propose a style augmentation strategy that changes the style of an image with remained structure information for augmentation, which can further improve the model's generalizability. %Comprehensive experiments on to prove that output softened anatomical features is critical to reconstruct fine-grained images as well as benefit for model generalization training. 
Quantitative experimental results on a multi-site fundus image dataset and a multi-domain NPC MRI dataset showed that our CDDSA outperformed several state-of-the-art multi-domain generalization methods. In the future, it is of interest to apply our CDDSA framework to other multi-domain medical image analysis tasks.



%% References


\bibliographystyle{model1-num-names}
\bibliography{refs.bib}






% \end{thebibliography}

\end{document}